\documentclass[10pt,twocolumn]{article}

\usepackage[utf8]{inputenc}
\usepackage[T1]{fontenc}
\usepackage{lmodern}
\usepackage{geometry}
\usepackage{graphicx}
\usepackage{amsmath,amssymb}
\usepackage{booktabs}
\usepackage[colorlinks=true, linkcolor=blue, citecolor=blue]{hyperref}
\usepackage[numbers]{natbib}
\usepackage{xcolor}
\usepackage{longtable}
\usepackage{multirow}
\usepackage{dblfloatfix}
\usepackage{cuted}
\usepackage{caption}
\usepackage{orcidlink}
\usepackage{float}
\title{\textbf{PulmoSight-XAI: An Explainable Multi-View Attention Ensemble with Gradient Boosting Meta-Learning for Multi-Label Chest X-Ray Classification}}

\author{
Moshiur Rahman$^{1}$\,\orcidlink{0009-0004-9663-8339}\quad 
Shafqat Alam$^{1}$\,\orcidlink{0009-0002-7374-3246}\quad
Tasnia Binte Mamun$^{1,*}$\,\orcidlink{0009-0001-0262-9545}
\\[1em]
$^{1}$Department of Biomedical Engineering\\
Bangladesh University of Engineering and Technology (BUET), 
Dhaka, Bangladesh
\\[0.5em]
$^{*}$Corresponding author
}

\date{}

\begin{document}

\twocolumn[
\maketitle
\begin{abstract}
Automated chest X-ray classification remains challenging due to severe class imbalance, complex co-occurring pathologies, and the loss of localized discriminative features in conventional deep learning architectures. To address these limitations, we propose an explainable hierarchical multi-view ensemble framework for robust classification of 14 thoracic pathologies. The proposed framework employs view-specific training by independently modeling frontal and lateral radiographs using an ensemble of five complementary convolutional neural networks. Instead of relying on conventional global average pooling, a multi-scale feature fusion strategy augmented with Convolutional Block Attention Modules (CBAM) is introduced to preserve fine-grained intermediate representations while emphasizing high-level pathology-specific semantic features. To mitigate both positive-negative imbalance and varying inter-class learning difficulty, the base models are optimized using a novel hybrid objective that combines Asymmetric Loss with Adaptive Focal Loss. Beyond conventional probability averaging, the framework incorporates a hierarchical meta-learning strategy in which test-time augmentation (TTA) predictions and cross-model uncertainty measures are integrated into Level-1 gradient-boosting meta-learners (XGBoost, LightGBM, and CatBoost), followed by Level-2 stacking with optimized alpha blending for final prediction. Evaluated on a large-scale CheXpert-style dataset, the proposed framework achieves state-of-the-art macro-average AUROC scores of 0.9319 for frontal and 0.9154 for lateral radiographs. Furthermore, comprehensive explainability analysis using seven complementary post-hoc attribution techniques demonstrates strong anatomical consistency and clinically meaningful decision localization. By integrating architectural diversity, multi-scale attention, hierarchical meta-learning, and rigorous explainability, the proposed framework provides a transparent, highly accurate, and clinically practical computer-aided diagnosis system for thoracic disease classification.
\end{abstract}
\vspace{0.5cm}
]

\section{Introduction}
A large proportion of hospital admissions and mortality in the world are due to thoracic diseases, with conditions including pneumonia, pleural effusion, and cardiomegaly causing a long-term strain on public-health systems \cite{ward_global_2024}. Chest X-ray (CXR) imaging is still the most important initial modality in the diagnosis of thoracic diseases due to benefits that include wide availability, low cost, and a lower radiation dose than computed tomography \cite{wang_chestx-ray8_2017}. The number of annual CXR studies, however, far exceeds the capacity of the available radiologist workforce. Countries, particularly in resource-limited contexts and rural settings, are facing a shortage of radiologists, resulting in delays in reporting and diagnostic mistakes \cite{mcisaac_global_2024, brady_error_2017}. All these pressures have sparked ongoing interest in deep-learning-based computer-aided diagnosis systems capable of reading CXR images reliably and being integrated into a clinical workflow \cite{litjens_survey_2017, johnson_mimic-cxr_2019}.

One of the most characteristic features of CXR interpretation, which makes it different from conventional image classification, is its multi-label nature: a single radiograph may contain a number of co-occurring findings, such as atelectasis with pleural effusion or cardiomegaly with pulmonary oedema \cite{chen_msa-net_2026}. This structure presents problems that single-label classifiers are not designed to handle, including severe class imbalance, complex inter-label dependencies, and high intra-class visual heterogeneity \cite{park_style-kd_2024}. Two benchmark datasets dominate the literature: the NIH ChestX-ray14 dataset, which is a collection of 112,120 frontal-view images across 14 pathology labels \cite{wang_chestx-ray8_2017}, and the Stanford CheXpert dataset, containing 224,316 multi-view radiographs taken at the front and side views \cite{irvin_chexpert_2019}.

Early deep learning approaches to multi-label CXR classification adapted ImageNet-pretrained CNN backbones. CheXNet established the foundation, showing that a 121-layer DenseNet fine-tuned on ChestX-ray14 could match radiologists on pneumonia detection, making DenseNet variants the default backbone for much of the subsequent work \cite{rajpurkar_chexnet_2017}. Later studies improved this paradigm by comparing different ResNet depths and transfer-learning strategies \cite{baltruschat_comparison_2018}, modelling inter-label dependencies using LSTM decoders \cite{yao_learning_2017}, and optimising the throughput--accuracy trade-off with GPU-oriented designs such as TResNet \cite{ridnik_tresnet_2020}. However, one aspect that always seems to be limited is the global average pooling (GAP) at the network head, which collapses the spatial structure into a single vector and discards the localized cues that are key to radiological reading. This motivated attention-based extensions, such as category-wise residual attention and channel--spatial modules like the Convolutional Block Attention Module (CBAM), which help to suppress background texture and highlight abnormal areas relevant to the disease \cite{guan_multi-label_2020, woo_cbam_2018}. The same emphasis on spatial detail produced hybrid CNN--Transformer models like HydraViT, which pairs a CNN spatial encoder with a Transformer context encoder to provide a balance between performance on rare and common classes \cite{ozturk_hydravit_2025}, and MXA, which augments EfficientViT by introducing region-of-interest pooling and knowledge distillation from a DenseNet teacher \cite{rand_beyond_2025}. More recent methods tend to be transformer-based, such as an optimised detection transformer (CD-DETR) \cite{yu_optimized_2025} and ConvNeXtV2-based frameworks with learnable global pooling \cite{xiong_multi-label_2025}, which have further improved the reported ROC-AUCs.

A parallel line of work aims at the extreme positive--negative imbalance of CXR datasets, in which the most common label (Support Devices) is two orders of magnitude more common than the rarest. Under this imbalance, standard binary cross-entropy (BCE) is dominated by the easy negatives and under-trains rare pathologies. Asymmetric Loss (ASL) takes care of this by decoupling the focusing exponents of positive and negative samples, down-weighting easy negatives so that gradient capacity aims at real cases of disease \cite{ben-baruch_asymmetric_2020}, and the method is now popular in CXR pipelines \cite{lu_cvtgnet_2024}. Related strategies encode clinical taxonomies through hierarchical cross-entropy losses \cite{asadi_clinically-inspired_2025} and re-weight pairwise label scores by classification difficulty using focal ranking losses \cite{hanif_enhancing_2025}. Complementary gains are seen at inference time: averaging predictions over deterministic test-time augmentations reduces aleatoric uncertainty, a principle first established for medical image segmentation \cite{wang_aleatoric_2019} and since extended to CXR classification. The CheXpert benchmark additionally raised the problem of uncertain labels, findings the automated report parser could not resolve as positive or negative for which Irvin et al. assessed imputation methods like U-Zeros and U-Ones \cite{irvin_chexpert_2019}.

Because independent prediction heads do not take into account the comorbidity relationships encoded in label co-occurrence, a large amount of research integrates graph convolutional networks (GCNs) into the classification pipeline \cite{lu_cvtgnet_2024}. Early formulations used the conditional probabilities of the training set to construct label co-occurrence adjacency matrices \cite{chen_label_2020}; this paradigm was then extended to consider in-batch images as graph nodes for cross-image semantic consistency \cite{chen_multi-label_2022} and to pair noise-robust Gaussian smoothing gates with a multimodal fusion head \cite{sun_multi-label_2026}. Subsequent variants include bipartite bridged GCNs \cite{wang_bb-gcn_2023}, similarity-graph distillation for missing-label inference \cite{ding_distilling_2025}, and multi-level pseudo-label consistency for single-positive-label supervision \cite{xiao_multi-label_2024}. The latest extensions include paired textual report embeddings to disentangle image--text features and apply category-disentangled causal learning, eliminating unwanted label correlations and enabling zero-shot generalisation \cite{mahapatra_multi-label_2025, li_multilabel_2026}.

In spite of this advancement, four constraints are seen in the literature repeatedly. First, the majority of the models are trained solely on frontal images or pool frontal and lateral views without specialisation, disregarding their complementary diagnostic value: frontal projections favour pneumothorax, whereas lateral projections favour pleural effusion, so the effect of per-view specialisation across all 14 CheXpert labels remains unquantified. Second, pipelines normally hold GAP at the deepest backbone stage, discarding the mid-level textural cues needed for subtle findings such as lung lesions or fractures; multi-scale feature fusion combined with attention is not well explored yet for this task. Third, conventional ensembles use a simple probability averaging that does not consider the differing confidence profiles and complementary error modes of the constituent architectures, and there has been limited use of structured meta-learning to explicitly model inter-architectural agreement. Fourth, a persistent long-tail deficit affects low-prevalence classes, an issue often camouflaged by macro-averaged metrics that hide weaknesses in each class.

To overcome these challenges, we suggest applying a multi-view ensemble with a hierarchical structure for the automated multi-label CXR classification problem. Here are some of our major contributions:
\begin{enumerate}
    \item We train five complementary CNN backbones (InceptionV3, ConvNeXtV2-Tiny, DenseNet201, EfficientNet-B5, and ResNeXt-101). These models are trained independently on frontal and lateral projections in order to take advantage of view-specific anatomical constraints.
    
    \item We avoid using native classification heads to extract hierarchical feature maps and instead refine the deep stages with CBAM, fusing them into a unified 512-dimensional representation containing both mid-level texture and high-level semantics.

    \item We present a combined Asymmetric and Adaptive Focal loss with label smoothing to jointly tackle global class imbalance and hard-to-detect rare pathologies.

    \item We design a stacked ensemble where Level-0 outputs, enhanced by test-time augmentation, consensus across models, and uncertainty estimates, are aggregated by Level-1 gradient-boosting learners (XGBoost, LightGBM, CatBoost) and a final Level-2 stacking / alpha-blending stage.

    \item The framework achieves macro-averaged AUROCs of 0.9319 (frontal) and 0.9154 (lateral) for 14 CheXpert pathologies, exceeding recent baselines, and we use seven post-hoc attribution methods to evaluate anatomical plausibility and describe systematic failure modes.

\end{enumerate}
The remainder of the paper is organised as follows. The dataset is described in Section~\ref{sec:dataset}. Section~\ref{sec:method} details the proposed methodology, covering image preprocessing, backbone architectures, attention and fusion mechanisms, the training strategy, and the hierarchical meta-learning ensemble. Section~\ref{sec:results} presents experimental results, ablation studies, cross-architecture comparisons, comparisons with previous work, and explainability analysis. Sections~\ref{sec:limitations} and~\ref{sec:conclusion} discuss limitations and end with suggestions for further research.

\section{Dataset}
\label{sec:dataset}
\begin{table*}[t]
  \centering
  \small
  \caption{Dataset statistics: split-level image distribution and per-pathology annotation totals for frontal and lateral views.}
  \label{tab:dataset_combined}
  \renewcommand{\arraystretch}{1.1}
  \setlength{\tabcolsep}{4pt}
  
  \begin{tabular}{l l ccc ccc}
    \toprule
    & & \multicolumn{3}{c}{\textbf{Frontal (AP + PA)}} & \multicolumn{3}{c}{\textbf{Lateral (Lat + LL)}} \\
    \cmidrule(lr){3-5}\cmidrule(lr){6-8}
    \textbf{Category} & \textbf{Split} & \textbf{Train} & \textbf{Val} & \textbf{Test} & \textbf{Train} & \textbf{Val} & \textbf{Test} \\
    \midrule
    \multirow{3}{*}{\textbf{Global}} 
      & Images    & 75,962 & 9,598 & 9,452 & 10,748 & 1,376 & 1,358 \\
      & Multi-Label (\%) & 52.29\% & 51.94\% & 52.39\% & 44.58\% & 42.51\% & 44.26\% \\
    \midrule
    \multirow{14}{*}{\textbf{Pathology}}
      & Atelectasis              & 27,745 & 3,468 & 3,468 & 3,145 & 394 & 393 \\
      & Cardiomegaly             & 25,181 & 3,147 & 3,148 & 2,803 & 350 & 351 \\
      & Consolidation            & 21,573 & 2,697 & 2,697 & 2,141 & 268 & 268 \\
      & Edema                    & 19,759 & 2,470 & 2,470 & 1,494 & 187 & 187 \\
      & Enl.\ Cardiomediastinum  & 27,153 & 3,394 & 3,394 & 2,900 & 362 & 363 \\
      & Fracture                 &  9,090 & 1,137 & 1,136 & 2,572 & 322 & 321 \\
      & Lung Lesion              &  9,050 & 1,131 & 1,132 &   835 & 105 & 104 \\
      & Lung Opacity             & 35,590 & 4,449 & 4,449 & 3,627 & 454 & 453 \\
      & No Finding               & 23,057 & 2,882 & 2,883 & 4,434 & 554 & 555 \\
      & Pleural Effusion         & 24,777 & 3,097 & 3,098 & 2,877 & 360 & 360 \\
      & Pleural Other            &  4,043 &   506 &   505 & 1,502 & 188 & 187 \\
      & Pneumonia                &  9,558 & 1,195 & 1,195 & 1,894 & 237 & 237 \\
      & Pneumothorax             &  6,322 &   790 &   791 &   668 &  83 &  84 \\
      & Support Devices          & 27,329 & 3,416 & 3,416 & 2,579 & 322 & 323 \\
    \bottomrule
  \end{tabular}
\end{table*}

We test the proposed model on Kaggle's Grand X-ray Slam Division-B dataset~\cite{grand-xray-slam-division-b}, a multi-label chest radiograph dataset annotated with 14 thoracic pathology labels using CheXpert-style conventions. The images cover two orientations: frontal projections (anteroposterior [AP] and posteroanterior [PA]) and lateral projections (standard lateral and left-lateral [LL]). There are 108,494 radiographs in the dataset in total---95,012 frontal and 13,482 lateral. We split this into training, validation, and test sets using an 80/10/10 stratified split (seed = 42). Multi-label stratification is applied so that across the three splits the per-class proportions are maintained.

The resulting splits and per-class frequencies make up Table~\ref{tab:dataset_combined} and make the label imbalance explicit. Lung Opacity is the most common class among frontal views (44,488), while Pleural Other (5,054) and Pneumothorax (7,903) are relatively uncommon. The lateral subset is smaller and more skewed, with Lung Lesion (1,044) and Pneumothorax (835) being the most underrepresented classes. Approximately 52\% of the frontal and 44\% of the lateral images carry more than one positive label, confirming the multi-label setting that the framework is designed to address.

\section{Methodology}
\label{sec:method}
This section introduces the proposed end-to-end pipeline for automated multi-label classification of CXR images. The framework consists of four steps: (1) image preprocessing and stochastic augmentation; (2) a set of diverse Level-0 base learners that combine multi-scale feature extraction with CBAM; (3) a training strategy built around an imbalance-aware hybrid loss and weight averaging; and (4) a meta-learning ensemble that aggregates the base learners using TTA and gradient boosting. One of the key aspects of the design is that the whole pipeline is duplicated, once for modelling frontal projections, and again for lateral projections so that each view is modelled by backbones specialised to its anatomy, rather than by a single network trained on pooled views.

\begin{figure*}[ht]
    \centering
    \includegraphics[width=\textwidth]{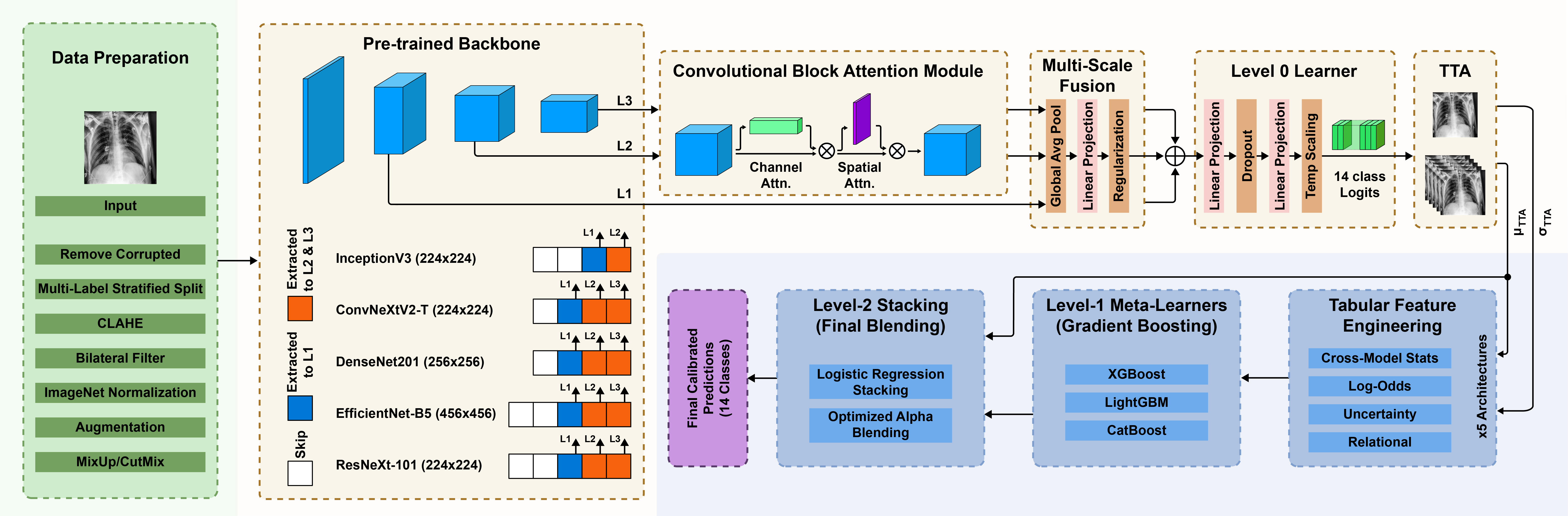}
    \caption{End-to-end overview of the proposed computational pipeline. The diagram illustrates the data preparation, the multi-scale feature extraction from the pre-trained backbones (stages $L_1, L_2, L_3$), the application of the Convolutional Block Attention Module (CBAM), and the subsequent Level-1 and Level-2 meta-learning ensemble processes.}
    \label{fig:architecture}
\end{figure*}

\subsection{Image Preprocessing and Augmentation}

We apply a two-step enhancement protocol to each raw radiograph. We first apply Contrast Limited Adaptive Histogram Equalization (CLAHE) with an $8 \times 8$ tile grid to amplify local contrast in low-visibility regions, then apply bilateral filtering to suppress the resulting noise while preserving anatomical edges. The enhanced images are resized to each model's native input resolution and normalised using ImageNet channel statistics.

During training we apply an extensive stochastic augmentation pipeline. Geometric and intensity-based transformations (random rotations, horizontal and vertical flips, affine shifts, elastic deformations) are combined with regularising perturbations including Coarse Dropout, grid distortion, and optical distortion. Beyond these, we incorporate MixUp~\cite{zhang_mixup_2018} and CutMix~\cite{yun_cutmix_2019}, applied to 40\% of training batches while the remaining 60\% use the standard pipeline. For MixUp, the mixing coefficient $\lambda$ is sampled from a Beta distribution; for CutMix, the label-combination ratio $\lambda$ is set by the area of the cropped region relative to the full image. Together these strategies expose each backbone to substantial appearance and label variation, which is particularly valuable for the rare classes that dominate the long tail of the dataset.

\subsection{Base Model Architectures (Level-0 Learners)}

We employ an ensemble of five CNN architectures as Level-0 base learners. As shown in Figure~\ref{fig:architecture}, the backbones are InceptionV3, ConvNeXtV2-Tiny, DenseNet201, EfficientNet-B5, and ResNeXt-101 (32$\times$8d), all initialised from ImageNet pre-trained weights. We select these networks for their architectural diversity: they differ in depth, receptive-field growth, and feature-reuse strategy, which in turn produces complementary error profiles that the meta-learning stage later exploits. Transfer learning from ImageNet additionally accelerates convergence and stabilises feature extraction in the low-annotation regime typical of medical imaging.

\subsubsection{Pre-trained Backbones and Hierarchical Feature Extraction}

Standard CNN classifiers terminate in a single GAP layer. While GAP preserves high-level semantics, it collapses the spatial structure that encodes the fine-grained, localized texture cues needed to detect subtle pulmonary findings. To recover these, we remove the native classification head of each backbone and instead read out intermediate 2D feature maps from several depths. Let the backbone be a feature extractor $f_{ext}(X; \theta)$, where $X \in \mathbb{R}^{3 \times H_0 \times W_0}$ is the input image and $\theta$ the network parameters. Processing $X$ through its sequential stages yields a set of hierarchical feature maps $\mathcal{F} = \{F_0, F_1, \dots, F_N\}$, from which we select a multi-scale subset $\{L_1, L_2, L_3\}$ (or $\{L_1, L_2\}$ for InceptionV3) spanning the mid- and deep-level stages. Each extracted stage $L_i$ is characterised by its channel capacity and spatial resolution, shown in~\ref{eq:1}:
\begin{equation}
\label{eq:1}
    L_i \in \mathbb{R}^{C_i \times H_i \times W_i}, \quad \text{for } i \in \{1, 2, 3\}
\end{equation}
where $C_i$ is the channel count and $H_i \times W_i$ the spatial resolution at stage $i$.

The routing of these stages is asymmetric by design, and this is a central element of the architecture. The shallow $L_1$ stage carries mid-level information and is forwarded directly to the fusion head, where its high spatial fidelity is preserved intact. The deeper $L_2$ and $L_3$ stages encode high-level semantics such as object parts and whole pathologies; because these maps also accumulate background activation, we route them through a CBAM before fusion so that channel and spatial attention can isolate the diagnostically relevant response. This separation is deliberate: applying attention to the deep stages sharpens semantic focus, while leaving $L_1$ unrefined retains the low-level detail that attention modules tend to smooth away. The motivation for reading out multiple scales at all is that no single depth captures the full range of thoracic abnormalities---fine, low-contrast findings such as Lung Lesion and Fracture surface at shallow stages, whereas large, high-contrast findings such as Cardiomegaly and Pleural Effusion are best represented deeper in the network. Of the five backbones, InceptionV3 uses a two-stage readout dictated by its stem-and-inception-block structure, while the remaining four use the full three-stage protocol; Table~\ref{tab:architectures} summarises the input resolutions, extracted stages, CBAM placement, and parameter counts for each.

\begin{table*}[ht]
\centering
\caption{Architectural specifications of the Level-0 base learners. The extracted stages correspond to the internal backbone layers used for multi-scale fusion. CBAM is selectively applied to the deepest extracted stages ($L_2$, $L_3$).}
\label{tab:architectures}
\begin{tabular}{lcccc}
\toprule
\textbf{Model Architecture} & \textbf{Input Size} & \textbf{Extracted Stages} & \textbf{CBAM Applied} & \textbf{Parameters} \\
\midrule
InceptionV3         & $224 \times 224$ & $L_1$ (Stage 2), $L_2$ (Stage 3)          & $L_2$       & $\sim$ 27 M \\
ConvNeXtV2-Tiny     & $224 \times 224$ & $L_1$ (Stage 1), $L_2$ (Stage 2), $L_3$ (Stage 3) & $L_2, L_3$  & $\sim$ 28 M \\
DenseNet201         & $256 \times 256$ & $L_1$ (Block 2), $L_2$ (Block 3), $L_3$ (Block 4) & $L_2, L_3$  & $\sim$ 20 M \\
EfficientNet-B5     & $456 \times 456$ & $L_1$ (Stage 2), $L_2$ (Stage 3), $L_3$ (Stage 4) & $L_2, L_3$  & $\sim$ 30 M \\
ResNeXt-101 (32x8d) & $224 \times 224$ & $L_1$ (Stage 2), $L_2$ (Stage 3), $L_3$ (Stage 4) & $L_2, L_3$  & $\sim$ 88 M \\
\bottomrule
\end{tabular}
\end{table*}

\subsubsection{Convolutional Block Attention Module (CBAM)}

We improve the deep $L_2$ and $L_3$ feature maps with a CBAM~\cite{woo_cbam_2018} that applies channel attention followed by spatial attention to emphasise clinically relevant responses and suppress background activation. For an intermediate feature map $\mathbf{F} \in \mathbb{R}^{C \times H \times W}$, the refinement proceeds as shown in~\ref{eq:2} and~\ref{eq:3}:
\begin{align}
    \label{eq:2}
    \mathbf{F}' &= \mathbf{M}_c(\mathbf{F}) \otimes \mathbf{F} \\
    \label{eq:3}
    \mathbf{F}'' &= \mathbf{M}_s(\mathbf{F}') \otimes \mathbf{F}'
\end{align}
where $\otimes$ denotes element-wise multiplication and $\mathbf{M}_c$, $\mathbf{M}_s$ are the channel and spatial attention maps. The channel attention map, which weights informative feature channels, is computed as:
\begin{equation}
    \mathbf{M}_c(\mathbf{F}) = \sigma \Big( \text{MLP}(\text{AvgPool}(\mathbf{F})) + \text{MLP}(\text{MaxPool}(\mathbf{F})) \Big)
\end{equation}
The spatial attention module then pools channel information into two 2D descriptors, concatenates them, and convolves the result with a $7 \times 7$ kernel:
\begin{equation}
    \mathbf{M}_s(\mathbf{F}') = \sigma \left( f^{7 \times 7} \left( [\text{AvgPool}(\mathbf{F}'); \text{MaxPool}(\mathbf{F}')] \right) \right)
\end{equation}
where $\sigma$ is the sigmoid activation and $f^{7 \times 7}$ a convolution with a $7 \times 7$ kernel. The refined output $\mathbf{F}''$ passed to the fusion head is therefore optimised along both the channel and spatial axes, which allows the deep stages to contribute sharp, pathology-focused semantics instead of a diffuse global activation.

\subsubsection{Multi-Scale Fusion Head}

The fusion head combines the raw mid-level stage ($L_1$) with the CBAM-refined deep stages ($L_2'', L_3''$) into a single 512-dimensional representation, so that the final descriptor carries both the low-level texture preserved at $L_1$ and the pathology-focused semantics sharpened by attention at the deeper stages. Each input feature map $F_k$ is first compressed by global average pooling and projected into a 512-dimensional latent space using a linear map $\phi_k$, a SiLU/GELU activation, and dropout:
\begin{equation}
    z_k = \phi_k\big( \text{AdaptiveAvgPool}(F_k) \big), \quad z_k \in \mathbb{R}^{512}
\end{equation}
The per-stage projections are then concatenated. The four three-stage backbones (ConvNeXtV2, DenseNet201, EfficientNet-B5, ResNeXt-101) produce a 1536-dimensional vector, while the two-stage InceptionV3 produces 1024:
\begin{equation}
    Z_{concat} = [z_1 \oplus z_2 \oplus z_3] \in \mathbb{R}^{1536} \quad (\text{or } \mathbb{R}^{1024} \text{ for InceptionV3})
\end{equation}
A terminal network $\psi$: linear projection, activation, and dropout, then synthesises this concatenation into the final feature vector:
\begin{equation}
    Z_{final} = \psi(Z_{concat}) \in \mathbb{R}^{512}
\end{equation}
$Z_{final}$ integrates early-stage texture with attention-refined semantics in a fixed-width representation, and serves as the shared bottleneck for both the classification head and the downstream meta-learners.

\subsubsection{Classification Head and Temperature Scaling}

The classification head maps $Z_{final}$ to the 14-dimensional label space. We apply dropout to $Z_{final}$ for regularisation, pass the result through a linear classifier, and convert the logits to probabilities with a temperature-scaled sigmoid. The scalar temperature $T$ is optimised jointly with the network weights, calibrating the entropy of the output distribution so that the predicted probabilities are neither systematically over- nor under-confident. Calibration matters here beyond the usual reasons: these probabilities are not only the final per-model output but also the raw material for the tabular features consumed by the Level-1 gradient-boosting meta-learners, which are sensitive to miscalibrated inputs.

\subsection{Training Strategy and Optimization}

The Grand X-ray Slam Division-B dataset exhibits extreme class imbalance and pronounced variation in diagnostic difficulty across its 14 labels. Standard objectives such as BCE treat all errors equally and therefore under-train the rare, difficult classes. We instead optimise the base learners using a combined loss that addresses two failure modes explicitly: global positive--negative imbalance and inter-class difficulty variance.

\subsubsection{Combined Hybrid Loss Formulation}

The training objective $\mathcal{L}_{Total}$ combines Asymmetric Loss (ASL), which targets global specificity under imbalance, with an Adaptive Focal Loss, which focuses on sensitivity on the most difficult classes. Each of the two terms has the same weight, The whole equation is stated in~\ref{eq:9}:
\begin{equation}
\label{eq:9}
    \mathcal{L}_{Total} = 0.5 \cdot \mathcal{L}_{ASL} + 0.5 \cdot \mathcal{L}_{AdaptiveFocal}
\end{equation}

\paragraph{Asymmetric Loss (ASL):} ASL~\cite{ben-baruch_asymmetric_2020} is a separation of the focusing applied to positive and negative samples. For a given image, let $y_c \in \{0, 1\}$ be the ground-truth label for the $c$-th pathology and $p_c \in [0, 1]$ the predicted probability. The class-wise loss is shown in~\ref{eq:10}:
\begin{equation}
\label{eq:10}
\begin{aligned}
\mathcal{L}_{ASL} = \frac{1}{K} \sum_{c=1}^{K} \Big(&
y_c (1-p_c)^{\gamma_{pos}} \log(p_c) \\
&+ (1-y_c) p_m^{\gamma_{neg}} \log(1-p_c)
\Big)
\end{aligned}
\end{equation}
The positive and negative focusing exponents are given as $\gamma_{pos}=1.0$ and $\gamma_{neg}=4.0$ respectively. The larger negative exponent, together with the margin-shifted probability $p_m$, suppresses the gradient contribution of easy negatives---the confidently correct background predictions that would otherwise dominate optimisation when there is extreme imbalance.

\paragraph{Adaptive Focal Loss:} To promote learning in the very hard-to-learn pathologies (Pneumothorax, Lung Lesion, Pleural Other, Fracture), we add a label-smoothed Adaptive Focal Loss. We smooth each target as $\tilde{y}_c = y_c(1 - \epsilon) + \frac{\epsilon}{2}$ with $\epsilon=0.1$, let $BCE_c$ be the binary cross-entropy between $p_c$ and $\tilde{y}_c$, and define $p_{t,c} = \exp(-BCE_c)$. The loss is stated in~\ref{eq:11}:
\begin{equation}
\label{eq:11}
    \mathcal{L}_{AdaptiveFocal} = \frac{1}{K} \sum_{c=1}^{K} \alpha (1 - p_{t,c})^{\Gamma_c} \cdot BCE_c
\end{equation}
using $\alpha = 0.25$ and a class-specific focal exponent $\Gamma_c$ assigned by difficulty: the identified ``hard'' classes have $\Gamma_c = 3.5$, and for other classes it is $\Gamma_{base} = 2.0$. Raising the exponent for hard classes focuses the gradient on instances that the model finds genuinely difficult, stabilising convergence despite the wide range of per-class difficulty. The two terms are complementary by construction: ASL controls the dominant easy-negative gradient across all classes, while the Adaptive Focal term redirects capacity toward the minority pathologies that ASL alone leaves under-served.

\subsubsection{Optimization and Convergence Strategies}

We accompany this objective with a set of optimisation techniques selected for efficiency and stable convergence. In order to make the memory cost of high-resolution inputs (up to $456 \times 456$) manageable, we train using Automatic Mixed Precision (FP16) and gradient accumulation, aggregating gradients over micro-batches to simulate larger effective batch sizes. Optimisation uses AdamW with a Cosine Annealing with Warm Restarts schedule. To make the model more robust on a noisy training landscape, we keep an exponential moving average (EMA, decay $\beta$) of the weights during training. After convergence, we apply Stochastic Weight Averaging (SWA) to settle into flatter, more generalisable minima.

\subsection{Advanced Ensemble Framework (Meta-Learning)}

The final stage combines the Level-0 base learners by using a multi-level ensemble. Rather than averaging their probabilities---which ignores the different confidence profiles and complementary error modes that motivated the architectural diversity in the first place---we encode those outputs into a structured tabular feature space and train Level-1 and Level-2 meta-learners to learn how to combine them.

\subsubsection{Test-Time Augmentation (TTA) and Feature Engineering}

We first use a deterministic TTA scheme to minimize the variance of the predictions and expose uncertainty. For each image, we evaluate five fixed, reproducible transformations---the original orientation, a horizontal flip, and subtle rotations of the object---and record the per-class mean prediction $\mu_{TTA}$ and standard deviation $\sigma_{TTA}$, the latter serving as an explicit uncertainty signal. These statistics are used in conjunction with the model outputs from the SWA to provide the basis of a tabular feature vector $X_{meta} \in \mathbb{R}^{N \times D}$.

We then add a number of feature blocks, each focusing on a particular aspect of ensemble behaviour. Cross-model statistics (mean, standard deviation, and range) measure the extent to which agreement is valid across all architectures, and the log-odds transform makes the probability space linear for the boosting stage. Per-sample scalar statistics: predictive entropy, class-wise probability rankings, and the number of positive predictions with high levels of confidence ($p > 0.5$) summarise the global nature of each image. Lastly, pairwise differences between the strongest base models, as well as the $\sigma_{TTA}$ uncertainty terms, flag cases where the architectures disagree. The result is a feature space that gives the meta-learners not only the point estimate, but an explicit description of the ensemble confidence and architectural divergence, which is exactly what simple averaging throws away.

\subsubsection{Level-1 Meta-Learning: Gradient Boosting}

The tabular feature matrix $X_{meta} \in \mathbb{R}^{N \times D}$, where $N$ is the number of samples and $D$ is the number of engineered features, is the input to the Level-1 layer. Because these features are tabular, we combine three gradient boosting algorithms with complementary inductive biases: XGBoost, LightGBM, and CatBoost. The algorithms are trained per class using a One-Versus-Rest (OVR) scheme, allowing a dedicated learner to focus on the signature of each of the 14 pathologies. To counter the class imbalance that persists after augmentation, every base learner $B_{c, algo}$ for pathology $c$ and algorithm $algo \in \{\text{XGB, LGB, Cat}\}$ is trained with a class-specific positive weight derived from the negative-to-positive ratio in the training set, shown at~\ref{eq:12}:
\begin{equation}
\label{eq:12}
w_{c, pos} = \frac{\sum_{n=1}^{N} (1 - y_{n,c})}{\max\left(\sum_{n=1}^{N} y_{n,c}, 1\right)} 
\end{equation}
where $y_{n,c} \in \{0, 1\}$ is the binary ground truth for sample $n$ and class $c$. Folding $w_{c, pos}$ into the objective scales up the gradient contribution of the scarce positive cases, countering the bias towards the majority (healthy) class that raw ensemble outputs inherit.

\subsubsection{Level-2 Ensemble: Stacking and Final Blending}

Level-2 combines the Level-1 predictions through two complementary strategies to get the final decision. The first one is logistic-regression stacking: for each pathology $c$, a linear meta-learner $\mathcal{M}_c$ learns the weighted combination of the XGBoost, LightGBM, and CatBoost outputs that best separates that class,
\begin{equation} \hat{y}_{c, stack} = \sigma \left( \beta_{0,c} + \sum_{i \in \{\text{XGB, LGB, Cat}\}} \beta_{i,c} \cdot \hat{p}_{c, i} \right) \end{equation}
where the coefficient $\beta_{i,c}$ indicates the relative reliability of the $i$-th booster and a regularisation parameter $C$ controls overfitting. The second strategy is an optimised alpha blend that performs linear interpolation between the boosted ensemble $\hat{y}_{booster}$ and the averaged Level-0 CNN outputs $\hat{y}_{cnn\_avg}$. A grid search for each class is used to identify the scalar $\alpha$ that maximises the macro-averaged AUROC:
\begin{equation} \begin{aligned} \hat{y}_{final} = \arg \max_{\alpha} \Big[ &\text{AUC}_{macro} \big( \alpha \cdot \hat{y}_{booster} \\ &+ (1 - \alpha) \cdot \hat{y}_{cnn\_avg} \big) \Big] \end{aligned} \end{equation}
with $\alpha \in [0, 0.55]$ sampled at 0.025 intervals. Optimising the value of $\alpha$ for each class yields a hybrid decision boundary: where boosting contributes little, the blend falls back on the more general CNN-ensembled representation; when a class has a unique statistical signature, the boosted features dominate. Thus it maintains sensitivity to the fine findings and protects against the overconfidence of any single deep model.

\subsection{Evaluation Framework}

Each radiograph can have multiple labels active at a time, so we assess per class and per aggregate. We report macro-averaged Accuracy, Sensitivity (Recall), Specificity, and $F_1$. For class $c \in \{1, \dots, 14\}$, let $TP_c$, $TN_c$, $FP_c$, and $FN_c$ be the true positives, true negatives, false positives, and false negatives at the operating threshold; these four measures use their usual definitions:
\begin{align}
    \text{Accuracy}_c &= \frac{TP_c + TN_c}{TP_c + TN_c + FP_c + FN_c} \\[4pt]
    \text{Sensitivity}_c &= \frac{TP_c}{TP_c + FN_c} \\[4pt]
    \text{Specificity}_c &= \frac{TN_c}{TN_c + FP_c} \\[4pt]
    F_{1, c} &= \frac{TP_c}{TP_c + \frac{1}{2}(FP_c + FN_c)}
\end{align}

Rather than a single universal threshold of 0.5 that is suboptimal for rare classes, we tune a per-class threshold $t_c$. For each class, the threshold $t_c$ is determined by doing a grid search over $[0.1, 0.9]$ in steps of $0.02$ using a validation set, picking the value that gives the highest value of Youden's Index (the average of sensitivity and specificity):
\begin{equation}
    t_c = \arg \max_{t \in \{0.1, \dots, 0.9\}} \left( \frac{\text{Sensitivity}_c(t) + \text{Specificity}_c(t)}{2} \right)
\end{equation}

We report global performance by macro-averaging each metric $\mathcal{M} \in \{\text{AUC}, \text{Sensitivity}, \text{Specificity}, F_1\}$ across all $K=14$ classes as $\mathcal{M}_{macro} = \frac{1}{K} \sum_{c=1}^{K} \mathcal{M}_c$, weighting every class equally regardless of frequency. AUROC is computed the same way, giving a threshold-independent measure of discrimination across the full operating range.

\section{Results and Discussion}
\label{sec:results}

Averaged over all backbones and classes, TTA gives only small directionally positive gains ($\Delta\text{AUROC} \approx 0.001$--$0.010$), its main benefit being variance reduction on minority and view-ambiguous classes rather than headline improvement (Table~\ref{tab:perclass_combined}). For classes which are already close to their discriminative ceiling, such as Enlarged Cardiomediastinum and Support Devices, the gains are negligible, confirming that marginal-variance averaging provides marginal gains once a backbone has saturated; as above, the biggest gains appear for minority classes in the lateral view. The isolated TTA regressions can be attributed to threshold sensitivity: averaging changes the output distribution slightly so that the output is shifted just enough to displace the Youden-optimal threshold calibrated using the validation set, which is the motivation behind the per-class recalibration performed at the meta-learning stage.

The per-class profiles in Table~\ref{tab:perclass_combined} show that there are complementary strengths that motivate the ensemble. ResNeXt101 ($\approx$88M parameters) has the best peak AUROCs on structurally complex pathologies, due to its grouped convolutions and high channel cardinality, but this capacity is reversed on lateral minority classes, where it shows the steepest sensitivity collapse (Lung Lesion 0.4904, Pneumothorax 0.3690). DenseNet201 is the most consistent on rare classes which is a consequence of dense skip pathways at low level which maintain boundary-discriminating properties that deeper networks suppress through progressive pooling. EfficientNet-B5's high-resolution input ($456 \times 456$) gives the highest frontal specificity on Lung Opacity (0.9393 with TTA) and competitive lateral Atelectasis (AUROC 0.9134). InceptionV3 records the lowest peak AUROCs but the highest TTA gains in absolute terms across minority and lateral classes, indicating its inception modules are more sensitive to positional perturbation, and thus benefit most from inference-time spatial diversity.

\begin{strip}
\makeatletter
\let\if@twocolumn\iffalse
\makeatother
\centering
{\scriptsize
\begin{longtable}{@{}p{1.9cm}p{1.7cm}|ccc|ccc||ccc|ccc@{}}
  \caption[Per-class metrics without and with TTA]{%
    Per-class classification metrics for all five individual backbone
    models evaluated on the held-out test set for \textbf{frontal} and
    \textbf{lateral} chest radiograph views, both \textbf{without} and
    \textbf{with test-time augmentation (TTA)}.
    All five architectures were trained on
    frontal and lateral images independently.
    TTA consists of $T{=}5$ deterministic transforms applied at
    inference: (i)~original image, (ii)~horizontal flip,
    (iii)~$+5^{\circ}$ rotation, (iv)~brightness/contrast enhancement,
    and (v)~centre-crop from a $1.05\times$ upscaled image.
  }
  \label{tab:perclass_combined}\\
  \toprule
  \multirow{3}{*}{\textbf{Pathology}} &
  \multirow{3}{*}{\textbf{Model}} &
  \multicolumn{6}{c||}{\textbf{Frontal}} &
  \multicolumn{6}{c}{\textbf{Lateral}} \\
  & & \multicolumn{3}{c}{\textbf{No TTA}} & \multicolumn{3}{c||}{\textbf{With TTA}} &
  \multicolumn{3}{c}{\textbf{No TTA}} & \multicolumn{3}{c}{\textbf{With TTA}} \\
  \cmidrule(lr){3-5}\cmidrule(lr){6-8}\cmidrule(lr){9-11}\cmidrule(lr){12-14}
  & & \textbf{AUC} & \textbf{Sen.} & \textbf{Spe.}
    & \textbf{AUC} & \textbf{Sen.} & \textbf{Spe.}
    & \textbf{AUC} & \textbf{Sen.} & \textbf{Spe.}
    & \textbf{AUC} & \textbf{Sen.} & \textbf{Spe.} \\
  \midrule
  \endfirsthead
  \multicolumn{14}{c}{\tablename\ \thetable{}~(\textit{continued})}\\[3pt]
  \toprule
  \multirow{3}{*}{\textbf{Pathology}} &
  \multirow{3}{*}{\textbf{Model}} &
  \multicolumn{6}{c||}{\textbf{Frontal}} &
  \multicolumn{6}{c}{\textbf{Lateral}} \\
  & & \multicolumn{3}{c}{\textbf{No TTA}} & \multicolumn{3}{c||}{\textbf{With TTA}} &
  \multicolumn{3}{c}{\textbf{No TTA}} & \multicolumn{3}{c}{\textbf{With TTA}} \\
  \cmidrule(lr){3-5}\cmidrule(lr){6-8}\cmidrule(lr){9-11}\cmidrule(lr){12-14}
  & & \textbf{AUC} & \textbf{Sen.} & \textbf{Spe.}
    & \textbf{AUC} & \textbf{Sen.} & \textbf{Spe.}
    & \textbf{AUC} & \textbf{Sen.} & \textbf{Spe.}
    & \textbf{AUC} & \textbf{Sen.} & \textbf{Spe.} \\
  \midrule
  \endhead
  \midrule\multicolumn{14}{r}{\textit{Continued on next page}}\\
  \endfoot
  \bottomrule
  \endlastfoot
  \multirow{5}{*}{\shortstack[l]{Atelectasis}} 
    & ConvNeXtV2-T & 0.9198 & 0.8287 & 0.8486 & 0.9205 & 0.8267 & 0.8516 & 0.9024 & 0.7328 & 0.8839 & 0.9033 & 0.7328 & 0.8891 \\
    & InceptionV3   & 0.9133 & 0.8062 & 0.8582 & 0.9175 & 0.8080 & 0.8656 & 0.9004 & 0.8193 & 0.8321 & 0.9071 & 0.8372 & 0.8301 \\
    & EfficientNet-B5 & 0.9203 & 0.8235 & 0.8636 & 0.9210 & 0.8241 & 0.8620 & 0.9132 & 0.7710 & 0.8819 & 0.9134 & 0.7812 & 0.8788 \\
    & DenseNet201   & 0.9073 & 0.7950 & 0.8694 & 0.9087 & 0.7956 & 0.8724 & 0.9087 & 0.8372 & 0.8321 & 0.9097 & 0.8372 & 0.8301 \\
    & ResNeXt101    & 0.9200 & 0.8155 & 0.8631 & 0.9210 & 0.8137 & 0.8656 & 0.9104 & 0.6947 & 0.9285 & 0.9155 & 0.6845 & 0.9399 \\
  \midrule
  \multirow{5}{*}{\shortstack[l]{Cardiomegaly}} 
    & ConvNeXtV2-T  & 0.9493 & 0.8987 & 0.8546 & 0.9495 & 0.8990 & 0.8552 & 0.8510 & 0.7265 & 0.7875 & 0.8581 & 0.7550 & 0.7855 \\
    & InceptionV3   & 0.9455 & 0.8841 & 0.8614 & 0.9466 & 0.8875 & 0.8618 & 0.8349 & 0.7550 & 0.7746 & 0.8480 & 0.7664 & 0.7716 \\
    & EfficientNet-B5 & 0.9472 & 0.8802 & 0.8672 & 0.9478 & 0.8799 & 0.8681 & 0.8384 & 0.6781 & 0.8590 & 0.8478 & 0.6809 & 0.8491 \\
    & DenseNet201   & 0.9431 & 0.8799 & 0.8631 & 0.9438 & 0.8793 & 0.8651 & 0.8434 & 0.8091 & 0.7378 & 0.8493 & 0.8348 & 0.7299 \\
    & ResNeXt101    & 0.9489 & 0.8815 & 0.8709 & 0.9500 & 0.8844 & 0.8715 & 0.8507 & 0.7236 & 0.8173 & 0.8601 & 0.7521 & 0.8054 \\
  \midrule
  \multirow{5}{*}{\shortstack[l]{Consolidation}} 
    & ConvNeXtV2-T  & 0.9394 & 0.8350 & 0.9052 & 0.9398 & 0.8339 & 0.9054 & 0.9149 & 0.8246 & 0.8798 & 0.9239 & 0.8470 & 0.8697 \\
    & InceptionV3   & 0.9333 & 0.8450 & 0.8847 & 0.9358 & 0.8517 & 0.8840 & 0.8979 & 0.7873 & 0.8615 & 0.9146 & 0.7910 & 0.8569 \\
    & EfficientNet-B5 & 0.9362 & 0.8432 & 0.8957 & 0.9373 & 0.8435 & 0.8948 & 0.8874 & 0.7985 & 0.8817 & 0.9064 & 0.8172 & 0.8761 \\
    & DenseNet201   & 0.9323 & 0.8584 & 0.8768 & 0.9329 & 0.8572 & 0.8778 & 0.9130 & 0.8209 & 0.8431 & 0.9216 & 0.8507 & 0.8257 \\
    & ResNeXt101    & 0.9364 & 0.8513 & 0.8932 & 0.9378 & 0.8539 & 0.8920 & 0.8815 & 0.6903 & 0.9330 & 0.9161 & 0.7239 & 0.9193 \\
  \midrule
  \multirow{5}{*}{\shortstack[l]{Edema}} 
    & ConvNeXtV2-T  & 0.9410 & 0.8429 & 0.8809 & 0.9413 & 0.8470 & 0.8821 & 0.8600 & 0.8235 & 0.7925 & 0.8834 & 0.8556 & 0.7797 \\
    & InceptionV3   & 0.9326 & 0.8907 & 0.8177 & 0.9350 & 0.9040 & 0.8149 & 0.8370 & 0.7005 & 0.8318 & 0.8607 & 0.7326 & 0.8155 \\
    & EfficientNet-B5 & 0.9380 & 0.8615 & 0.8645 & 0.9390 & 0.8615 & 0.8645 & 0.8599 & 0.8182 & 0.8061 & 0.8718 & 0.8342 & 0.7788 \\
    & DenseNet201   & 0.9346 & 0.8713 & 0.8502 & 0.9353 & 0.8628 & 0.8528 & 0.8672 & 0.8021 & 0.8079 & 0.8812 & 0.8182 & 0.8010 \\
    & ResNeXt101    & 0.9388 & 0.8684 & 0.8526 & 0.9398 & 0.8745 & 0.8541 & 0.8502 & 0.6257 & 0.8950 & 0.8862 & 0.6952 & 0.8736 \\
  \midrule
  \multirow{5}{*}{\shortstack[l]{Enl.\ Cardio-\\mediastinum}} 
    & ConvNeXtV2-T  & 0.9788 & 0.9455 & 0.9055 & 0.9789 & 0.9455 & 0.9047 & 0.9082 & 0.8650 & 0.8281 & 0.9142 & 0.8760 & 0.8201 \\
    & InceptionV3   & 0.9767 & 0.9372 & 0.9134 & 0.9773 & 0.9364 & 0.9114 & 0.8987 & 0.8512 & 0.8161 & 0.9081 & 0.8733 & 0.8191 \\
    & EfficientNet-B5 & 0.9785 & 0.9526 & 0.9026 & 0.9789 & 0.9511 & 0.9011 & 0.9051 & 0.8650 & 0.8251 & 0.9098 & 0.8650 & 0.8131 \\
    & DenseNet201   & 0.9770 & 0.9425 & 0.9044 & 0.9777 & 0.9446 & 0.9044 & 0.9065 & 0.8650 & 0.8131 & 0.9135 & 0.8843 & 0.8090 \\
    & ResNeXt101    & 0.9794 & 0.9523 & 0.9010 & 0.9798 & 0.9558 & 0.8990 & 0.9117 & 0.7631 & 0.8884 & 0.9165 & 0.7686 & 0.8945 \\
  \midrule
  \pagebreak
  \midrule
  \multirow{5}{*}{\shortstack[l]{Fracture}} 
    & ConvNeXtV2-T  & 0.9343 & 0.8637 & 0.8536 & 0.9351 & 0.8637 & 0.8505 & 0.9002 & 0.7819 & 0.8920 & 0.9033 & 0.7913 & 0.8843 \\
    & InceptionV3   & 0.9209 & 0.8734 & 0.8231 & 0.9250 & 0.8848 & 0.8172 & 0.8950 & 0.8069 & 0.8284 & 0.9012 & 0.8255 & 0.8110 \\
    & EfficientNet-B5 & 0.9332 & 0.8602 & 0.8765 & 0.9348 & 0.8610 & 0.8754 & 0.8923 & 0.8069 & 0.8833 & 0.8969 & 0.8162 & 0.8746 \\
    & DenseNet201   & 0.9276 & 0.8883 & 0.8142 & 0.9291 & 0.8945 & 0.8072 & 0.8910 & 0.7819 & 0.8650 & 0.8973 & 0.7757 & 0.8640 \\
    & ResNeXt101    & 0.9384 & 0.8777 & 0.8637 & 0.9397 & 0.8821 & 0.8637 & 0.8956 & 0.7072 & 0.9219 & 0.9059 & 0.7321 & 0.9257 \\
  \midrule
  \multirow{5}{*}{\shortstack[l]{Lung Lesion}} 
    & ConvNeXtV2-T  & 0.8588 & 0.7659 & 0.7905 & 0.8602 & 0.7694 & 0.7974 & 0.8672 & 0.7981 & 0.8110 & 0.8730 & 0.8173 & 0.7951 \\
    & InceptionV3   & 0.8461 & 0.7244 & 0.8053 & 0.8539 & 0.7473 & 0.8005 & 0.8251 & 0.6923 & 0.8006 & 0.8777 & 0.8269 & 0.7839 \\
    & EfficientNet-B5 & 0.8615 & 0.7553 & 0.8108 & 0.8634 & 0.7580 & 0.8085 & 0.7983 & 0.6827 & 0.9027 & 0.8341 & 0.7212 & 0.8931 \\
    & DenseNet201   & 0.8448 & 0.6749 & 0.8389 & 0.8466 & 0.6802 & 0.8441 & 0.8242 & 0.7500 & 0.7974 & 0.8351 & 0.8077 & 0.7616 \\
    & ResNeXt101    & 0.8651 & 0.7367 & 0.8376 & 0.8675 & 0.7367 & 0.8425 & 0.7902 & 0.4904 & 0.9625 & 0.8514 & 0.5096 & 0.9522 \\
  \midrule
  \multirow{5}{*}{\shortstack[l]{Lung Opacity}} 
    & ConvNeXtV2-T  & 0.9116 & 0.7811 & 0.8985 & 0.9119 & 0.7802 & 0.9001 & 0.9065 & 0.7020 & 0.9293 & 0.9114 & 0.7130 & 0.9304 \\
    & InceptionV3   & 0.9015 & 0.7755 & 0.8897 & 0.9046 & 0.7822 & 0.8908 & 0.9027 & 0.8344 & 0.8088 & 0.9098 & 0.8565 & 0.8044 \\
    & EfficientNet-B5 & 0.9087 & 0.7390 & 0.9379 & 0.9093 & 0.7363 & 0.9393 & 0.9155 & 0.8256 & 0.8740 & 0.9164 & 0.8212 & 0.8740 \\
    & DenseNet201   & 0.9041 & 0.7532 & 0.9168 & 0.9056 & 0.7557 & 0.9192 & 0.9107 & 0.8389 & 0.8331 & 0.9111 & 0.8411 & 0.8188 \\
    & ResNeXt101    & 0.9084 & 0.7815 & 0.8950 & 0.9095 & 0.7820 & 0.8969 & 0.9156 & 0.7748 & 0.8939 & 0.9211 & 0.7770 & 0.9028 \\
  \midrule
  \multirow{5}{*}{\shortstack[l]{No Finding}} 
    & ConvNeXtV2-T  & 0.9194 & 0.8789 & 0.8031 & 0.9198 & 0.8796 & 0.8011 & 0.9156 & 0.8360 & 0.8531 & 0.9195 & 0.8396 & 0.8593 \\
    & InceptionV3   & 0.9104 & 0.8727 & 0.7903 & 0.9139 & 0.8740 & 0.7984 & 0.9115 & 0.8523 & 0.8244 & 0.9188 & 0.8559 & 0.8207 \\
    & EfficientNet-B5 & 0.9225 & 0.8713 & 0.8211 & 0.9235 & 0.8727 & 0.8211 & 0.9242 & 0.8937 & 0.8257 & 0.9265 & 0.8883 & 0.8381 \\
    & DenseNet201   & 0.9107 & 0.8799 & 0.7832 & 0.9112 & 0.8810 & 0.7817 & 0.9218 & 0.8342 & 0.8543 & 0.9252 & 0.8234 & 0.8643 \\
    & ResNeXt101    & 0.9237 & 0.8786 & 0.8150 & 0.9246 & 0.8789 & 0.8151 & 0.9213 & 0.9045 & 0.7945 & 0.9286 & 0.9189 & 0.7870 \\
  \midrule
  \multirow{5}{*}{\shortstack[l]{Pleural Effusion}} 
    & ConvNeXtV2-T  & 0.9197 & 0.8631 & 0.8201 & 0.9208 & 0.8654 & 0.8166 & 0.9367 & 0.8444 & 0.8908 & 0.9402 & 0.8611 & 0.8838 \\
    & InceptionV3   & 0.9131 & 0.8337 & 0.8360 & 0.9160 & 0.8382 & 0.8400 & 0.9349 & 0.8889 & 0.8447 & 0.9386 & 0.9028 & 0.8367 \\
    & EfficientNet-B5 & 0.9178 & 0.8450 & 0.8411 & 0.9185 & 0.8469 & 0.8396 & 0.9228 & 0.8333 & 0.9038 & 0.9338 & 0.8472 & 0.8928 \\
    & DenseNet201   & 0.9146 & 0.8499 & 0.8212 & 0.9161 & 0.8495 & 0.8225 & 0.9313 & 0.9139 & 0.8126 & 0.9343 & 0.9250 & 0.7996 \\
    & ResNeXt101    & 0.9191 & 0.8405 & 0.8485 & 0.9205 & 0.8405 & 0.8505 & 0.9235 & 0.7806 & 0.9198 & 0.9365 & 0.8139 & 0.9178 \\
  \midrule
  \multirow{5}{*}{\shortstack[l]{Pleural Other}} 
    & ConvNeXtV2-T  & 0.8734 & 0.7806 & 0.8164 & 0.8731 & 0.7866 & 0.8163 & 0.8645 & 0.7487 & 0.8582 & 0.8686 & 0.7647 & 0.8480 \\
    & InceptionV3   & 0.8506 & 0.7431 & 0.8393 & 0.8545 & 0.7391 & 0.8390 & 0.8590 & 0.7701 & 0.7950 & 0.8760 & 0.8075 & 0.7857 \\
    & EfficientNet-B5 & 0.8552 & 0.7767 & 0.8238 & 0.8620 & 0.7826 & 0.8237 & 0.8432 & 0.7807 & 0.8412 & 0.8439 & 0.7647 & 0.8155 \\
    & DenseNet201   & 0.8495 & 0.7589 & 0.7914 & 0.8542 & 0.7648 & 0.7882 & 0.8855 & 0.8075 & 0.7959 & 0.8858 & 0.8182 & 0.8002 \\
    & ResNeXt101    & 0.8740 & 0.7925 & 0.8105 & 0.8777 & 0.8004 & 0.8098 & 0.8630 & 0.6738 & 0.9266 & 0.8827 & 0.7059 & 0.9223 \\
  \midrule
  \multirow{5}{*}{\shortstack[l]{Pneumonia}} 
    & ConvNeXtV2-T  & 0.9079 & 0.8510 & 0.8121 & 0.9088 & 0.8527 & 0.8119 & 0.8519 & 0.7679 & 0.7574 & 0.8562 & 0.7975 & 0.7449 \\
    & InceptionV3   & 0.8888 & 0.7916 & 0.8323 & 0.8957 & 0.8042 & 0.8320 & 0.8326 & 0.8017 & 0.7145 & 0.8499 & 0.8439 & 0.6789 \\
    & EfficientNet-B5 & 0.8942 & 0.8192 & 0.8225 & 0.8976 & 0.8276 & 0.8224 & 0.8113 & 0.7384 & 0.8055 & 0.8228 & 0.7679 & 0.7939 \\
    & DenseNet201   & 0.9022 & 0.8226 & 0.8218 & 0.9039 & 0.8243 & 0.8233 & 0.8414 & 0.6456 & 0.8546 & 0.8442 & 0.6835 & 0.8421 \\
    & ResNeXt101    & 0.9025 & 0.8335 & 0.8044 & 0.9063 & 0.8427 & 0.8017 & 0.8191 & 0.6034 & 0.8760 & 0.8541 & 0.6582 & 0.8671 \\
  \midrule
  \multirow{5}{*}{\shortstack[l]{Pneumothorax}} 
    & ConvNeXtV2-T  & 0.9037 & 0.7481 & 0.8877 & 0.9031 & 0.7481 & 0.8887 & 0.8213 & 0.7381 & 0.8234 & 0.8412 & 0.7976 & 0.8077 \\
    & InceptionV3   & 0.8836 & 0.8063 & 0.8137 & 0.8914 & 0.8177 & 0.8034 & 0.8324 & 0.7143 & 0.8352 & 0.8640 & 0.7976 & 0.8250 \\
    & EfficientNet-B5 & 0.9214 & 0.8316 & 0.8717 & 0.9245 & 0.8304 & 0.8687 & 0.8514 & 0.6548 & 0.9356 & 0.8738 & 0.7024 & 0.9286 \\
    & DenseNet201   & 0.8781 & 0.7899 & 0.8137 & 0.8821 & 0.8013 & 0.8111 & 0.8119 & 0.6429 & 0.8768 & 0.8326 & 0.6905 & 0.8673 \\
    & ResNeXt101    & 0.9256 & 0.8658 & 0.8429 & 0.9299 & 0.8684 & 0.8433 & 0.7576 & 0.3690 & 0.9827 & 0.8592 & 0.3929 & 0.9749 \\
  \midrule
  \multirow{5}{*}{\shortstack[l]{Support Devices}} 
    & ConvNeXtV2-T  & 0.9703 & 0.9333 & 0.8919 & 0.9702 & 0.9374 & 0.8911 & 0.9043 & 0.8576 & 0.7816 & 0.9088 & 0.8762 & 0.7585 \\
    & InceptionV3   & 0.9651 & 0.9221 & 0.8903 & 0.9664 & 0.9245 & 0.8921 & 0.8970 & 0.8080 & 0.8164 & 0.9064 & 0.8235 & 0.8184 \\
    & EfficientNet-B5 & 0.9745 & 0.9552 & 0.8937 & 0.9747 & 0.9587 & 0.8937 & 0.9103 & 0.8297 & 0.8899 & 0.9159 & 0.8297 & 0.8870 \\
    & DenseNet201   & 0.9556 & 0.9180 & 0.8551 & 0.9553 & 0.9145 & 0.8553 & 0.8975 & 0.7585 & 0.8609 & 0.9050 & 0.7802 & 0.8609 \\
    & ResNeXt101    & 0.9755 & 0.9529 & 0.8987 & 0.9758 & 0.9526 & 0.8995 & 0.9096 & 0.7028 & 0.9266 & 0.9253 & 0.7214 & 0.9314 \\
  \bottomrule
\end{longtable}
}
\makeatletter
\let\if@twocolumn\iftrue 
\makeatother
\end{strip}

  These four profiles: peak capacity (ResNeXt101), rare-class sensitivity (DenseNet201), resolution-driven specificity (EfficientNet-B5), and augmentation responsiveness (InceptionV3) are designed to complement each other and explain the consistent macro-average gains the ensemble achieves over every individual backbone on both views.

\begin{table*}[!t]
\centering
\caption{Best ensemble performance on the test set.}
\label{tab:best_ensemble}
\begin{tabular}{lcccccc}
\toprule
& \multicolumn{3}{c}{\textbf{Frontal}} & \multicolumn{3}{c}{\textbf{Lateral}} \\
\cmidrule(lr){2-4} \cmidrule(lr){5-7}
\textbf{Pathology} & \textbf{AUC} & \textbf{Sen.} & \textbf{Spe.} & \textbf{AUC} & \textbf{Sen.} & \textbf{Spe.} \\
\midrule
Atelectasis       & 0.924 & 0.834 & 0.856 & 0.927 & 0.886 & 0.820 \\
Cardiomegaly      & 0.952 & 0.906 & 0.853 & 0.871 & 0.849 & 0.739 \\
Consolidation     & 0.941 & 0.847 & 0.903 & 0.935 & 0.855 & 0.871 \\
Edema             & 0.943 & 0.886 & 0.857 & 0.901 & 0.893 & 0.779 \\
Enl. Cardiomed.   & \textbf{0.981} & 0.954 & 0.906 & 0.923 & 0.854 & 0.862 \\
Fracture          & 0.944 & 0.902 & 0.852 & 0.922 & 0.801 & \textbf{0.910} \\
Lung Lesion       & 0.878 & 0.759 & 0.835 & 0.897 & 0.740 & 0.876 \\
Lung Opacity      & 0.914 & 0.814 & 0.873 & 0.929 & 0.846 & 0.875 \\
No Finding        & 0.928 & 0.878 & 0.829 & 0.936 & 0.885 & 0.852 \\
Pleural Effusion  & 0.924 & 0.841 & 0.854 & \textbf{0.951} & \textbf{0.919} & 0.867 \\
Pleural Other     & 0.891 & 0.652 & \textbf{0.924} & 0.915 & 0.866 & 0.810 \\
Pneumonia         & 0.915 & 0.842 & 0.830 & 0.877 & 0.722 & 0.857 \\
Pneumothorax      & 0.937 & 0.808 & 0.907 & 0.898 & 0.798 & 0.875 \\
Support Devices   & 0.977 & \textbf{0.959} & 0.902 & 0.934 & 0.845 & 0.872 \\
\midrule
\textbf{Macro Avg.} & \textbf{0.932} & \textbf{0.849} & \textbf{0.870} & \textbf{0.915} & \textbf{0.840} & \textbf{0.847} \\
\bottomrule
\end{tabular}
\end{table*}

\begin{table*}[htbp]
\centering
\caption{Macro-averaged performance summary: Individual backbone baseline versus the final ensemble.}
\label{tab:performance_summary}

\begin{tabular}{lcccccc}
\toprule
& \multicolumn{3}{c}{\textbf{Frontal}} & \multicolumn{3}{c}{\textbf{Lateral}} \\
\cmidrule(lr){2-4} \cmidrule(lr){5-7}
\textbf{Model Configuration} & \textbf{AUC} & \textbf{Sens.} & \textbf{Spec.} & \textbf{AUC} & \textbf{Sens.} & \textbf{Spec.} \\
\midrule
ConvNeXtV2-T & 0.9234 & \textbf{0.8415} & 0.8655 & 0.8942 & 0.7712 & 0.8546 \\
InceptionV3 & 0.9156 & 0.8247 & 0.8532 & 0.8878 & 0.7834 & 0.8213 \\
EfficientNet-B5 & 0.9218 & 0.8389 & \textbf{0.8682} & 0.8953 & 0.7688 & 0.8710 \\
DenseNet201 & 0.9102 & 0.8146 & 0.8519 & \textbf{0.8995} & \textbf{0.7915} & 0.8354 \\
ResNeXt101 & \textbf{0.9258} & 0.8423 & 0.8647 & 0.8847 & 0.7024 & \textbf{0.8942} \\
\midrule
\textbf{Final Ensemble} & \textbf{0.9319} & \textbf{0.8486} & \textbf{0.8702} & \textbf{0.9154} & \textbf{0.8398} & \textbf{0.8474} \\
\bottomrule
\end{tabular}

\end{table*}

The most consistent and strongest performers for all backbones and for the ensemble are the pathologies with large-scale and high-contrast morphology. These are stated at Tables~\ref{tab:best_ensemble} and~\ref{tab:perclass_combined}. Enlarged Cardiomediastinum has the maximal frontal AUROC in the whole evaluation (0.981 at the ensemble level, while those of the backbone were 0.9794 and 0.9788 for ResNeXt101 and ConvNeXtV2-Tiny without TTA). This reflects the posteroanterior projection providing the best view of the cardiac silhouette, with CBAM spatial attention suppressing background parenchymal texture. Support Devices follow closely (frontal AUROC 0.977, sensitivity 0.959) due to the discrete rectilinear shape of the implanted devices, and Cardiomegaly reaches the second-highest frontal AUROC (0.952), indicating that the ensemble combines forecasts that are already good at the backbone level.

\begin{table*}[!b]
\centering
\caption{Per-class AUROC comparison of the proposed final ensemble against seven published baselines evaluated on CheXpert 1.0.}
\label{table:comparison}
\small
\begin{tabular}{lcccccccc}
\toprule
\textbf{Pathology} & \textbf{Ours} & \textbf{SSGE \cite{chen_multi-label_2022}} & \textbf{GCF-Net \cite{sun_multi-label_2026}} & \textbf{U-Zeros \cite{irvin_chexpert_2019}} & \textbf{cheXGCN \cite{rajpurkar_chexnet_2017}} & \textbf{CvTGNet \cite{yu_optimized_2025}} & \textbf{MXA \cite{rand_beyond_2025}} \\ \midrule
Atelectasis & \textbf{0.925} & 0.728 & 0.721 & 0.730 & 0.736 & 0.731 & 0.770 \\
Cardiomegaly & \textbf{0.941} & 0.887 & 0.879 & 0.876 & 0.876 & 0.900 & 0.690 \\
Consolidation & \textbf{0.946} & 0.728 & 0.745 & 0.775 & 0.784 & 0.784 & 0.810 \\
Edema & \textbf{0.938} & 0.890 & 0.875 & 0.881 & 0.886 & 0.864 & 0.530 \\
Enl. Cardiomed. & \textbf{0.973} & 0.678 & 0.673 & 0.691 & 0.697 & 0.793 & 0.870 \\
Fracture & \textbf{0.941} & 0.829 & 0.813 & 0.807 & 0.833 & 0.853 & --- \\
Lung Lesion & \textbf{0.880} & 0.795 & 0.806 & 0.766 & 0.768 & 0.842 & 0.320 \\
Lung Opacity & \textbf{0.916} & 0.809 & 0.763 & 0.807 & 0.822 & 0.736 & 0.850 \\
No Finding & \textbf{0.929} & 0.889 & 0.869 & 0.811 & 0.825 & 0.802 & 0.820 \\
Pleural Effusion & \textbf{0.927} & 0.915 & 0.907 & 0.900 & 0.907 & 0.881 & 0.910 \\
Pleural Other & 0.894 & 0.813 & 0.790 & 0.815 & 0.835 & \textbf{0.925} & 0.750 \\
Pneumonia & \textbf{0.910} & 0.799 & 0.797 & 0.804 & 0.810 & 0.868 & 0.650 \\
Pneumothorax & \textbf{0.933} & 0.913 & 0.894 & 0.902 & 0.917 & 0.889 & 0.790 \\
Support Devices & \textbf{0.971} & 0.927 & 0.907 & 0.893 & 0.899 & 0.860 & 0.820 \\ \bottomrule
\end{tabular}
\end{table*}

For pathologies controlled by anatomical geometry and fluid movement caused by gravity, a significant asymmetry in projection is observed. Pleural Effusion is always more vigorous laterally at both the backbone and ensemble levels, reaching the best lateral AUROC for the evaluation (0.951, sensitivity 0.919) compared with a frontal AUROC of 0.924. Pneumothorax shows the opposite and most severe pattern: the ensemble frontal AUROC of 0.937 drops to 0.898 laterally, a gap that widens sharply at the backbone level, where ResNeXt101 falls from an AUROC of 0.9256 (sensitivity 0.8658) in front position to an AUROC of only 0.7576 and sensitivity of 0.3690 laterally. The cause is mechanical: bilateral superimposition of the lung fields in the lateral projection makes the unilateral hyperlucency hard to see, which is the most important diagnostic feature. TTA somewhat lessens this by broadening the sampled spatial context, but it is not able to recover the lateral sensitivity of ResNeXt101, which points to an architectural rather than stochastic deficit. Consolidation and Edema remain competitive across views---Consolidation at 0.935 laterally versus 0.941 frontally, Edema at 0.901 versus 0.943---with TTA's highest sensitivity gain for Edema under InceptionV3.

The greatest clinical discrepancy is at the extremes of the label distribution, at both the backbone and ensemble levels. Lung Lesion has the lowest frontal ensemble AUROC (0.878), roughly 0.09--0.12 points lower than the top classes; the lateral sensitivity of ResNeXt101 without TTA is 0.4904, which is the lowest value in the per-class evaluation, and although the ensemble recovers the lateral AUROC to 0.897, the class is still quite challenging. TTA again helps most here, improving the lateral sensitivity of InceptionV3 from 0.6923 to 0.8269. Pleural Other and Pneumonia share this profile reflecting severe imbalance compounded by visual overlap with co-occurring Consolidation and Edema.

The macro-averaged AUROCs of the proposed hierarchical ensemble are 0.9319 (frontal) and 0.9154 (lateral), exceeding the 0.89--0.91 range typically reported for 14-class thoracic classification, shown at Table~\ref{tab:performance_summary}. This gain comes from two design decisions: the multi-scale fusion paradigm, which keeps high-frequency textural detail that is lost through global average pooling, and the structured meta-learning layer, which models inter-architectural consensus rather than averaging probabilities.

When looking at the five Level-0 backbones, ResNeXt101 has a higher score on the frontal AUROC of 0.9258, while DenseNet201 (0.8995) and EfficientNet-B5 (0.8953) lead laterally. Transitions from any single backbone to the Level-2 ensemble improve both sensitivity and specificity on both views, thus confirming that the architectural diversity of the constituents is a functional advantage rather than redundancy. The LR-stacking configuration is used for the frontal ensemble and the optimised blend for the lateral ensemble; these macro-level figures (Table~\ref{tab:performance_summary}) consolidate, at a glance, the granular per-class and per-backbone trends discussed above.

\subsection{Contextual Positioning Relative to Prior Work}

Table~\ref{table:comparison} compares the AUROC of our proposed hierarchical ensemble with seven baselines from the thoracic imaging literature. Before interpreting these numbers, it is important to make clear that our results are based on the competition data from the Grand-X-ray Slam Division-B, which is a proprietary split with CheXpert-like labelling conventions, while all baselines are reported on the public CheXpert v1.0 benchmark. Currently, no published study has reported any results on the Grand-X-ray Slam dataset, so it is not possible to compare these with other reported results. The differences in prevalence of the labels, protocol of the annotations, and acquisition characteristics between the two datasets precludes the use of the baseline scores reported for SSGE \cite{chen_multi-label_2022}, GCF-Net \cite{sun_multi-label_2026}, U-Zeros \cite{irvin_chexpert_2019}, and the others as a controlled benchmark against our system; rather, they are to be regarded as contextual references. No significance testing can be performed between these two evaluation settings, because the test sets used to calculate the AUROC estimates under these two settings are not identical, and hence the sampling distributions of the AUROC estimates are not the same.

With that caveat in place, the reported results are still useful as a sanity check of overall system behaviour. Our ensemble outperforms the reported baseline AUROCs for most of the 14 classes of pathology, including those that are usually regarded as being difficult due to low-contrast features or semantic overlap, including Atelectasis (0.9247, versus baseline range 0.72--0.77) and Consolidation (0.9405, versus baseline range 0.73--0.81). Even though it is difficult to attribute this gap solely to architectural choices given the dataset mismatch, the direction and strength of the trend is generally in line with the design intuition of our multi-scale feature fusion paradigm: the backbone stages capture high-frequency textural details at three different scales, and then progressively refine them using CBAM attention modules, while most of the baseline architectures only use a single GAP layer. This serves as supporting context, not a proof point---the architectural claim is supported more directly by the within-dataset ablations presented elsewhere in this work.

Another key difference of our system is the view-stratified training methodology. Our Level-0 models are trained independently on the frontal and lateral images, enabling each of the backbones to focus on the anatomical constraints characteristic to its view of the image, such as the cardiac silhouette in the frontal view or the posterior costophrenic sulcus in the lateral view. A weighted AUROC formula is used to create a single diagnostic metric from per-view predictions, shown in~\ref{eq:20}:

\begin{equation}
\label{eq:20}
\text{Weighted AUC} = \frac{(N_F \times \text{AUC}_F) + (N_L \times \text{AUC}_L)}{N_F + N_L}
\end{equation}

where $N_F$ and $N_L$ are the number of frontal and lateral images in the test set, respectively, and $\text{AUC}_F$ and $\text{AUC}_L$ are the macro-averaged AUROCs for the frontal and lateral images, respectively.

This specialization is most evident, within our own dataset, in pathologies with high geometric clarity. Enlarged Cardiomediastinum and Support Devices reach AUROCs of 0.9733 and 0.9713, respectively, well above the CheXpert-reported baseline range, reflecting the ensemble's effective utilization of the high-contrast features of implanted devices and the clear morphological changes at the mediastinal border. Our Pneumonia AUROC (0.9103) also exceeds the range of baselines reported (0.79--0.86); although the difference is likely due to the dataset mismatch noted above, another plausible explanation is our hybrid loss function, which uses an Adaptive Focal Loss to focus the gradient budget on classes which are often masked by co-occurring Edema or Opacity findings.

\subsection{Explainability Analysis}
\label{subsec:xai}

\begin{figure*}[!htbp]
    \centering
    \includegraphics[width=\textwidth, height=0.8\textheight, keepaspectratio]{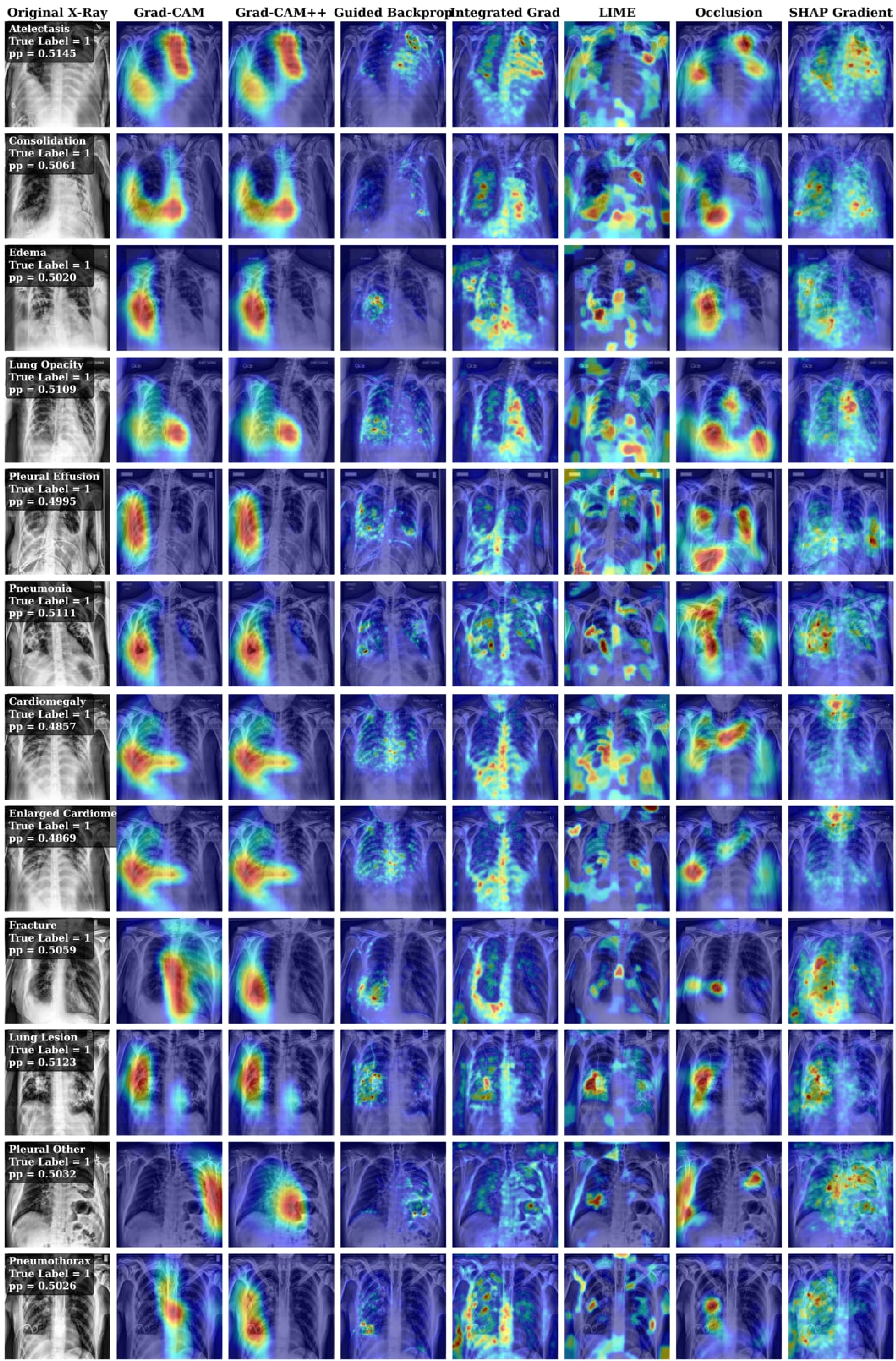}
    \caption{XAI heatmap visualizations of frontal chest X-rays obtained using DenseNet-201. Each row represents a case with a confirmed positive label from the following classes: Atelectasis, Cardiomegaly, Consolidation, Edema, Enlarged Cardiomediastinum, Fracture, Lung Lesion, Lung Opacity, Pleural Effusion, Pleural Other, Pneumonia, and Pneumothorax. Column~1 displays the original CLAHE-enhanced image, while Columns~2--8 show attribution maps produced by Grad-CAM, Grad-CAM++, Guided Backpropagation, Integrated Gradients, LIME, Occlusion, and SHAP Gradient, respectively.}
    \label{fig:xai_frontal}
\end{figure*}

\begin{figure*}[!htbp]
    \centering
    \includegraphics[width=\textwidth, height=0.8\textheight, keepaspectratio]{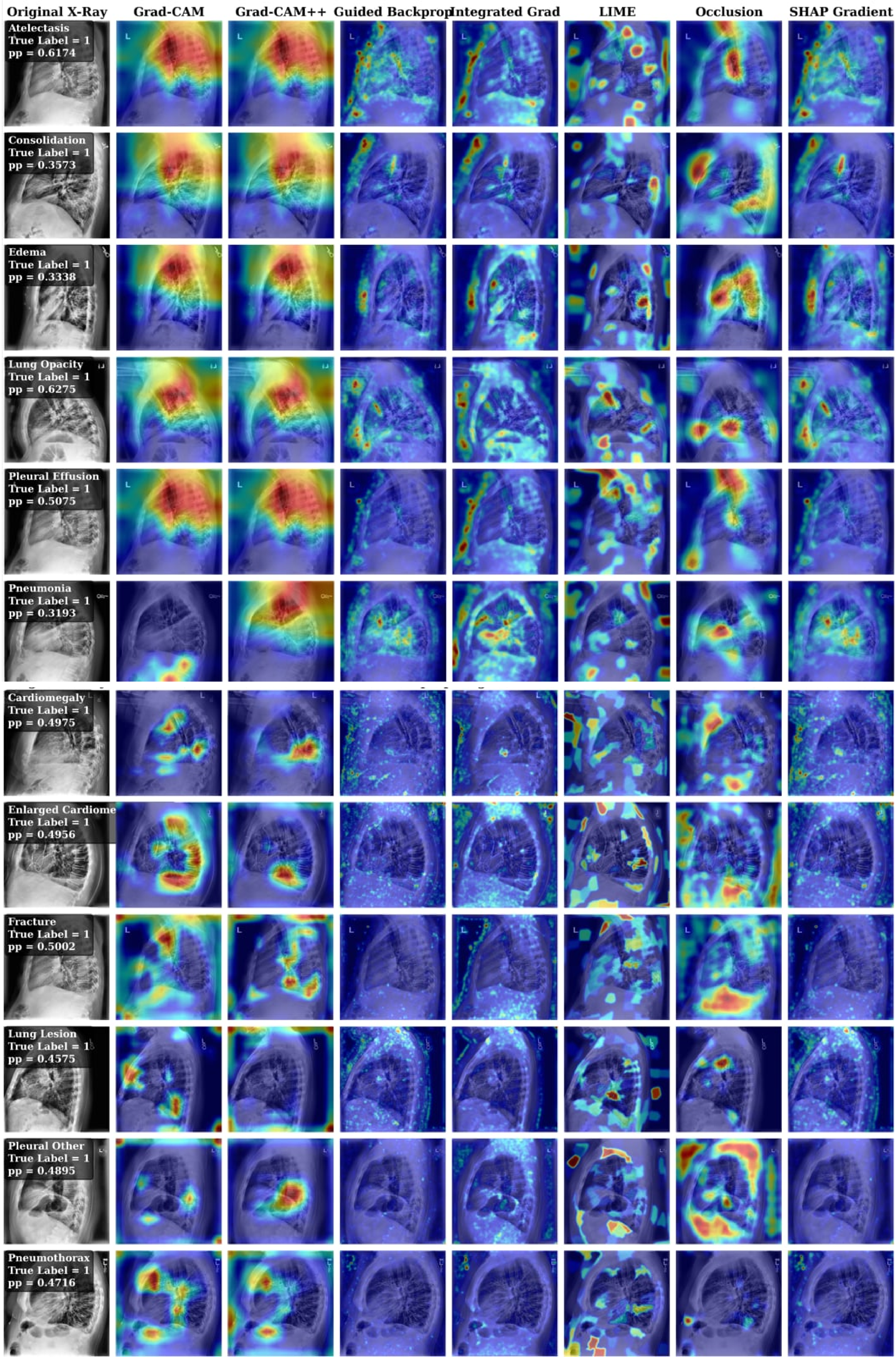}
    \caption{XAI heatmap visualisations of lateral chest X-rays. Each row represents a case with a confirmed positive label from the following classes: Atelectasis, Cardiomegaly, Consolidation, Edema, Enlarged Cardiomediastinum, Fracture, Lung Lesion, Lung Opacity, Pleural Effusion, Pleural Other, Pneumonia, and Pneumothorax. The first six rows are generated using ConvNeXtV2-Tiny, and the last six rows using EfficientNet-B5. Column~1 shows the original CLAHE-enhanced image, while Columns~2--8 present attribution maps produced by Grad-CAM, Grad-CAM++, Guided Backpropagation, Integrated Gradients, LIME, Occlusion, and SHAP Gradient, respectively.}
    \label{fig:xai_lateral}
\end{figure*}

To test the learned representations and to localise systematic failure modes, we applied seven complementary post-hoc attribution methods---Grad-CAM \cite{selvaraju_grad-cam_2016}, Grad-CAM++ \cite{chattopadhay_grad-cam_2018}, Guided Backpropagation \cite{springenberg_striving_2015}, Integrated Gradients \cite{sundararajan_axiomatic_2017}, LIME \cite{ribeiro_why_2016}, Occlusion \cite{zeiler_visualizing_2013}, and SHAP Gradient \cite{lundberg_unified_2017}---to confirmed positive test instances across all 14 classes in both projections. Combined, these maps show that predictions are based on anatomically relevant areas instead of dataset-specific confounders, and they show the circumstances under which discriminative capacity degrades.

For geometrically different afflictions, the attributions correlate to radiologic landmarks. In frontal views (Figure~\ref{fig:xai_frontal}), gradient-based methods localise to the cardiac silhouette for Cardiomegaly, the pleural margin for Pneumothorax, and the body of implanted hardware for Support Devices, confirming that the CBAM-refined features capture the high-contrast markers responsible for the good performance on these classes (AUROC $\approx$ 0.97). Lateral views (Figure~\ref{fig:xai_lateral}) are complementary sources of information for gravity-dependent findings; for example, Pleural Effusion localises to the costophrenic angle. This view-specific behaviour supports the multi-scale fusion design and takes into account the benefits of view-stratified training on conditions that are identifiable in the lateral projection but obscured in the frontal.

Diffuse and long-tail pathologies are more difficult to localize. For Atelectasis, gradient methods produce wider activation fields than those of LIME and Occlusion, which is consistent with signal dilution in the model. Saliency leaking into areas of co-occurring pathologies is seen in long-tail classes like Lung Lesion and Pneumonia, indicating that Asymmetric Loss does not fully neutralise the effect of label imbalance on the decision boundary. The model reliably takes advantage of high-frequency features for structural targets, but needs additional guidance such as weakly-supervised spatial losses to enhance discrimination of vague and underrepresented discoveries, a direction we will see again in Section~\ref{sec:limitations}.

\section{Limitations and Future Directions}
\label{sec:limitations}
While the results are impressive, there are a number of limitations that warrant mention. The evaluation is done on one proprietary CheXpert-style competition dataset, which makes it hard to compare with public benchmarks and may limit generalisation to out-of-distribution clinical settings and varying acquisition protocols \cite{stacke_measuring_2021}. The multi-label stratified splitting was used to split the dataset, but the co-occurrence of labels does not necessarily represent real-world prevalence in heterogeneous multi-institutional cohorts, so further external validation on other datasets like MIMIC-CXR or PadChest is needed \cite{bustos_padchest_2020}. The current loss formulation does not fully solve the problem of class imbalance, as can be seen in the decrease in sensitivity of long-tail pathologies like Lung Lesion, Pleural Other, and Pneumonia. Future work should focus on diffusion-based synthetic radiograph generation and few-shot and self-supervised pretraining methods appropriate for medical imaging with low annotations \cite{chambon_roentgen_2022, azizi_big_2021}.

Multiple large backbones and meta-learners make the framework computationally complex, which prevents the implementation of the framework in resource-limited environments. Structured pruning, quantization-aware training, and knowledge distillation to a lightweight student network are viable approaches for model compression \cite{han_deep_2015}. Furthermore, uncertainty estimates in the current system are missing; while test-time augmentation will lower variance, integrating Bayesian approximations or conformal prediction would make clinically meaningful uncertainty bounds possible \cite{angelopoulos_conformal_2022}. Adding weakly supervised spatial losses or bounding-box supervision could further enhance localization of diffuse findings, as is reflected by saliency effects seen in the explainability analysis. The current post-hoc attribution pipeline does not quantitatively assess the quality of localization (e.g., IoU or pointing game scores) with respect to ground-truth masks, however; future research should incorporate such measurements \cite{wang_chestx-ray8_2017}. Lastly, prospective multi-site validation and fairness auditing across demographic subgroups are still crucial, due to the well-known domain shift and bias issues in medical imaging \cite{seyyed-kalantari_underdiagnosis_2021}.

\section{Conclusion}
\label{sec:conclusion}
The proposed multi-view ensemble framework is a hierarchical approach that classifies chest X-rays automatically into 14 thoracic pathology classes. The proposed system is trained with five complementary CNN backbones on both frontal and lateral projections, which is followed by the integration of multi-scale feature fusion, refinement by CBAM, and the addition of test-time augmentation uncertainty estimates and cross-model consensus features into the structured gradient-boosting meta-learning layer, to obtain a macro-averaged AUROC of 0.9319 on frontal and 0.9154 on lateral radiographs, which significantly outperforms the 14-class classifiers with the AUROC range of 0.89--0.91 reported in previous works. The view-stratified training paradigm turns out to be particularly useful for geometrically asymmetric pathologies like Pleural Effusion and Pneumothorax, and the hybrid Asymmetric and Adaptive Focal loss effectively enhances the sensitivity on the rare and confusing classes. Using seven complementary post-hoc attribution analyses, the anatomical plausibility for structurally distinct findings is confirmed, further highlighting diffuse pathologies as the main target for future refinement of the model. The overall results show that architectural diversity, structured meta-learning, and view-specific specialisation are complementary and individually essential components of a high-performance thoracic CAD system, and are hoped to provide a good foundation for clinically deployable automated radiograph interpretation.

\bibliographystyle{ieeetr}
\bibliography{NewBib}

@article{xiong_multi-label_2025,
	title = {Multi-{Label} {Disease} {Detection} in {Chest} {X}-{Ray} {Imaging} {Using} a {Fine}-{Tuned} {ConvNeXtV2} with a {Customized} {Classifier}},
	volume = {12},
	issn = {2227-9709},
	url = {https://www.mdpi.com/2227-9709/12/3/80},
	doi = {10.3390/informatics12030080},
	language = {en},
	number = {3},
	urldate = {2026-03-26},
	journal = {Informatics},
	author = {Xiong, Kangzhe and Tu, Yuyun and Rao, Xinping and Zou, Xiang and Du, Yingkui},
	month = aug,
	year = {2025},
	pages = {80},
}

@article{hanif_enhancing_2025,
	title = {Enhancing {Multi}-{Label} {Chest} {X}-{Ray} {Classification} {Using} an {Improved} {Ranking} {Loss}},
	volume = {12},
	issn = {2306-5354},
	url = {https://www.mdpi.com/2306-5354/12/6/593},
	doi = {10.3390/bioengineering12060593},
	language = {en},
	number = {6},
	urldate = {2026-03-26},
	journal = {Bioengineering},
	author = {Hanif, Muhammad Shehzad and Bilal, Muhammad and Alsaggaf, Abdullah H. and Al-Saggaf, Ubaid M.},
	month = may,
	year = {2025},
	pages = {593},
}

@article{yu_optimized_2025,
	title = {An optimized transformer model for efficient detection of thoracic diseases in chest {X}-rays with multi-scale feature fusion},
	volume = {20},
	issn = {1932-6203},
	url = {https://dx.plos.org/10.1371/journal.pone.0323239},
	doi = {10.1371/journal.pone.0323239},
	language = {en},
	number = {5},
	urldate = {2026-03-26},
	journal = {PLOS One},
	author = {Yu, Shasha and Zhou, Peng},
	editor = {Shaikh,, Asadullah},
	month = may,
	year = {2025},
	pages = {e0323239},
}

@article{li_multilabel_2026,
	title = {Multilabel {Chest} {X}-{Ray} {Image} {Classification} via {Category} {Disentangled} {Causal} {Learning}},
	volume = {7},
	copyright = {https://ieeexplore.ieee.org/Xplorehelp/downloads/license-information/IEEE.html},
	issn = {2691-4581},
	url = {https://ieeexplore.ieee.org/document/11089950/},
	doi = {10.1109/TAI.2025.3591094},
	number = {2},
	urldate = {2026-03-26},
	journal = {IEEE Transactions on Artificial Intelligence},
	author = {Li, Qiang and Liu, Mengdi and Chang, Rihao and Nie, Weizhi and Bai, Shaojin and Liu, Anan},
	month = feb,
	year = {2026},
	pages = {1048--1061},
}

@article{xiao_multi-label_2024,
	title = {Multi-{Label} {Chest} {X}-{Ray} {Image} {Classification} {With} {Single} {Positive} {Labels}},
	volume = {43},
	copyright = {https://ieeexplore.ieee.org/Xplorehelp/downloads/license-information/IEEE.html},
	issn = {0278-0062, 1558-254X},
	url = {https://ieeexplore.ieee.org/document/10579876/},
	doi = {10.1109/TMI.2024.3421644},
	number = {12},
	urldate = {2026-03-26},
	journal = {IEEE Transactions on Medical Imaging},
	author = {Xiao, Jiayin and Li, Si and Lin, Tongxu and Zhu, Jian and Yuan, Xiaochen and Feng, David Dagan and Sheng, Bin},
	month = dec,
	year = {2024},
	pages = {4404--4418},
}

@article{mahapatra_multi-label_2025,
	title = {Multi-{Label} {Generalized} {Zero} {Shot} {Chest} {X}-{Ray} {Classification} by {Combining} {Image}-{Text} {Information} {With} {Feature} {Disentanglement}},
	volume = {44},
	copyright = {https://creativecommons.org/licenses/by-nc-nd/4.0/},
	issn = {0278-0062, 1558-254X},
	url = {https://ieeexplore.ieee.org/document/10601163/},
	doi = {10.1109/TMI.2024.3429471},
	number = {1},
	urldate = {2026-03-26},
	journal = {IEEE Transactions on Medical Imaging},
	author = {Mahapatra, Dwarikanath and Jimeno Yepes, Antonio and Bozorgtabar, Behzad and Roy, Sudipta and Ge, Zongyuan and Reyes, Mauricio},
	month = jan,
	year = {2025},
	pages = {31--43},
}

@misc{asadi_clinically-inspired_2025,
	title = {Clinically-{Inspired} {Hierarchical} {Multi}-{Label} {Classification} of {Chest} {X}-rays with a {Penalty}-{Based} {Loss} {Function}},
	url = {http://arxiv.org/abs/2502.03591},
	doi = {10.48550/arXiv.2502.03591},
	urldate = {2026-03-26},
	publisher = {arXiv},
	author = {Asadi, Mehrdad and Sodok{\'e}, Komi and Gerard, Ian J. and Kersten-Oertel, Marta},
	month = feb,
	year = {2025},
	note = {arXiv:2502.03591 [cs]},
}

@misc{rajpurkar_chexnet_2017,
	title = {{CheXNet}: {Radiologist}-{Level} {Pneumonia} {Detection} on {Chest} {X}-{Rays} with {Deep} {Learning}},
	copyright = {arXiv.org perpetual, non-exclusive license},
	shorttitle = {{CheXNet}},
	url = {https://arxiv.org/abs/1711.05225},
	doi = {10.48550/ARXIV.1711.05225},
	urldate = {2026-03-27},
	publisher = {arXiv},
	author = {Rajpurkar, Pranav and Irvin, Jeremy and Zhu, Kaylie and Yang, Brandon and Mehta, Hershel and Duan, Tony and Ding, Daisy and Bagul, Aarti and Langlotz, Curtis and Shpanskaya, Katie and Lungren, Matthew P. and Ng, Andrew Y.},
	year = {2017},
}

@article{wang_chestx-ray8_2017,
	title = {{ChestX}-ray8: {Hospital}-scale {Chest} {X}-ray {Database} and {Benchmarks} on {Weakly}-{Supervised} {Classification} and {Localization} of {Common} {Thorax} {Diseases}},
    journal = {arXiv preprint arXiv:1705.02315},
	copyright = {arXiv.org perpetual, non-exclusive license},
	shorttitle = {{ChestX}-ray8},
	url = {https://arxiv.org/abs/1705.02315},
	doi = {10.48550/ARXIV.1705.02315},
	urldate = {2026-03-27},
	publisher = {arXiv},
	author = {Wang, Xiaosong and Peng, Yifan and Lu, Le and Lu, Zhiyong and Bagheri, Mohammadhadi and Summers, Ronald M.},
	year = {2017},
}

@misc{yao_learning_2017,
	title = {Learning to diagnose from scratch by exploiting dependencies among labels},
	copyright = {arXiv.org perpetual, non-exclusive license},
	url = {https://arxiv.org/abs/1710.10501},
	doi = {10.48550/ARXIV.1710.10501},
	urldate = {2026-03-27},
	publisher = {arXiv},
	author = {Yao, Li and Poblenz, Eric and Dagunts, Dmitry and Covington, Ben and Bernard, Devon and Lyman, Kevin},
	year = {2017},
}

@misc{baltruschat_comparison_2018,
	title = {Comparison of {Deep} {Learning} {Approaches} for {Multi}-{Label} {Chest} {X}-{Ray} {Classification}},
	copyright = {arXiv.org perpetual, non-exclusive license},
	url = {https://arxiv.org/abs/1803.02315},
	doi = {10.48550/ARXIV.1803.02315},
	urldate = {2026-03-27},
	publisher = {arXiv},
	author = {Baltruschat, Ivo M. and Nickisch, Hannes and Grass, Michael and Knopp, Tobias and Saalbach, Axel},
	year = {2018},
}

@misc{woo_cbam_2018,
	title = {{CBAM}: {Convolutional} {Block} {Attention} {Module}},
	copyright = {arXiv.org perpetual, non-exclusive license},
	shorttitle = {{CBAM}},
	url = {https://arxiv.org/abs/1807.06521},
	doi = {10.48550/ARXIV.1807.06521},
	urldate = {2026-03-27},
	publisher = {arXiv},
	author = {Woo, Sanghyun and Park, Jongchan and Lee, Joon-Young and Kweon, In So},
	year = {2018},
}

@misc{ben-baruch_asymmetric_2020,
	title = {Asymmetric {Loss} {For} {Multi}-{Label} {Classification}},
	copyright = {arXiv.org perpetual, non-exclusive license},
	url = {https://arxiv.org/abs/2009.14119},
	doi = {10.48550/ARXIV.2009.14119},
	urldate = {2026-03-27},
	publisher = {arXiv},
	author = {Ben-Baruch, Emanuel and Ridnik, Tal and Zamir, Nadav and Noy, Asaf and Friedman, Itamar and Protter, Matan and Zelnik-Manor, Lihi},
	year = {2020},
}

@misc{ridnik_tresnet_2020,
	title = {{TResNet}: {High} {Performance} {GPU}-{Dedicated} {Architecture}},
	copyright = {arXiv.org perpetual, non-exclusive license},
	shorttitle = {{TResNet}},
	url = {https://arxiv.org/abs/2003.13630},
	doi = {10.48550/ARXIV.2003.13630},
	urldate = {2026-03-27},
	publisher = {arXiv},
	author = {Ridnik, Tal and Lawen, Hussam and Noy, Asaf and Baruch, Emanuel Ben and Sharir, Gilad and Friedman, Itamar},
	year = {2020},
}

@article{wang_aleatoric_2019,
	title = {Aleatoric uncertainty estimation with test-time augmentation for medical image segmentation with convolutional neural networks},
	volume = {338},
	issn = {09252312},
	url = {https://linkinghub.elsevier.com/retrieve/pii/S0925231219301961},
	doi = {10.1016/j.neucom.2019.01.103},
	language = {en},
	urldate = {2026-03-27},
	journal = {Neurocomputing},
	author = {Wang, Guotai and Li, Wenqi and Aertsen, Michael and Deprest, Jan and Ourselin, S{\'e}bastien and Vercauteren, Tom},
	month = apr,
	year = {2019},
	pages = {34--45},
}

@misc{irvin_chexpert_2019,
	title = {{CheXpert}: {A} {Large} {Chest} {Radiograph} {Dataset} with {Uncertainty} {Labels} and {Expert} {Comparison}},
	copyright = {Creative Commons Attribution 4.0 International},
	shorttitle = {{CheXpert}},
	url = {https://arxiv.org/abs/1901.07031},
	doi = {10.48550/ARXIV.1901.07031},
	urldate = {2026-03-27},
	publisher = {arXiv},
	author = {Irvin, Jeremy and Rajpurkar, Pranav and Ko, Michael and Yu, Yifan and Ciurea-Ilcus, Silviana and Chute, Chris and Marklund, Henrik and Haghgoo, Behzad and Ball, Robyn and Shpanskaya, Katie and Seekins, Jayne and Mong, David A. and Halabi, Safwan S. and Sandberg, Jesse K. and Jones, Ricky and Larson, David B. and Langlotz, Curtis P. and Patel, Bhavik N. and Lungren, Matthew P. and Ng, Andrew Y.},
	year = {2019},
}

@article{sun_multi-label_2026,
	title = {Multi-label chest {X}-ray image classification based on graph convolutional networks and multi-modal fusion},
	volume = {119},
	issn = {17468094},
	url = {https://linkinghub.elsevier.com/retrieve/pii/S1746809426004842},
	doi = {10.1016/j.bspc.2026.109930},
	language = {en},
	urldate = {2026-04-14},
	journal = {Biomedical Signal Processing and Control},
	author = {Sun, Junding and Hu, Jiayao and Wu, Xiaosheng and Xu, Zhaozhao and Wang, Yuwen and Zhang, Yudong},
	month = jun,
	year = {2026},
	pages = {109930},
}

@misc{rand_beyond_2025,
	title = {Beyond {Conventional} {Transformers}: {The} {Medical} {X}-ray {Attention} ({MXA}) {Block} for {Improved} {Multi}-{Label} {Diagnosis} {Using} {Knowledge} {Distillation}},
	shorttitle = {Beyond {Conventional} {Transformers}},
	url = {http://arxiv.org/abs/2504.02277},
	doi = {10.48550/arXiv.2504.02277},
	urldate = {2026-04-14},
	publisher = {arXiv},
	author = {Rand, Amit and Ibrahim, Hadi},
	month = may,
	year = {2025},
	note = {arXiv:2504.02277 [cs]},
}

@article{chen_multi-label_2022,
	title = {Multi-{Label} {Chest} {X}-{Ray} {Image} {Classification} via {Semantic} {Similarity} {Graph} {Embedding}},
	volume = {32},
	copyright = {https://ieeexplore.ieee.org/Xplorehelp/downloads/license-information/IEEE.html},
	issn = {1051-8215, 1558-2205},
	url = {https://ieeexplore.ieee.org/document/9430552/},
	doi = {10.1109/TCSVT.2021.3079900},
	number = {4},
	urldate = {2026-04-14},
	journal = {IEEE Transactions on Circuits and Systems for Video Technology},
	author = {Chen, Bingzhi and Zhang, Zheng and Li, Yingjian and Lu, Guangming and Zhang, David},
	month = apr,
	year = {2022},
	pages = {2455--2468},
}

@inproceedings{lu_cvtgnet_2024,
	address = {Ischia Italy},
	title = {{CvTGNet}: {A} {Novel} {Framework} for {Chest} {X}-{Ray} {Multi}-label {Classification}},
	isbn = {979-8-4007-0597-7},
	shorttitle = {{CvTGNet}},
	url = {https://dl.acm.org/doi/10.1145/3649153.3649216},
	doi = {10.1145/3649153.3649216},
	language = {en},
	urldate = {2026-04-14},
	booktitle = {Proceedings of the 21st {ACM} {International} {Conference} on {Computing} {Frontiers}},
	publisher = {ACM},
	author = {Lu, Yu and Hu, Yating and Li, Leya and Xu, Zhanpeng and Liu, Hongwei and Liang, Huanwen and Fu, Xianghua},
	month = may,
	year = {2024},
	pages = {12--20},
}

@article{chen_label_2020,
	title = {Label {Co}-{Occurrence} {Learning} {With} {Graph} {Convolutional} {Networks} for {Multi}-{Label} {Chest} {X}-{Ray} {Image} {Classification}},
	volume = {24},
	copyright = {https://ieeexplore.ieee.org/Xplorehelp/downloads/license-information/IEEE.html},
	issn = {2168-2194, 2168-2208},
	url = {https://ieeexplore.ieee.org/document/8961143/},
	doi = {10.1109/JBHI.2020.2967084},
	number = {8},
	urldate = {2026-04-14},
	journal = {IEEE Journal of Biomedical and Health Informatics},
	author = {Chen, Bingzhi and Li, Jinxing and Lu, Guangming and Yu, Hongbing and Zhang, David},
	month = aug,
	year = {2020},
	pages = {2292--2302},
}

@article{mcisaac_global_2024,
	title = {Global {Strategy} on {Human} {Resources} for {Health}: {Workforce} 2030-{A} {Five}-{Year} {Check}-{In}},
	volume = {22},
	issn = {1478-4491},
	shorttitle = {Global {Strategy} on {Human} {Resources} for {Health}},
	doi = {10.1186/s12960-024-00940-x},
	language = {eng},
	number = {1},
	journal = {Human Resources for Health},
	author = {McIsaac, Michelle and Buchan, James and Abu-Agla, Ayat and Kawar, Rania and Campbell, James},
	month = oct,
	year = {2024},
	pages = {68},
}

@article{brady_error_2017,
	title = {Error and discrepancy in radiology: inevitable or avoidable?},
	volume = {8},
	issn = {1869-4101},
	shorttitle = {Error and discrepancy in radiology},
	doi = {10.1007/s13244-016-0534-1},
	language = {eng},
	number = {1},
	journal = {Insights into Imaging},
	author = {Brady, Adrian P.},
	month = feb,
	year = {2017},
	pages = {171--182},
}

@article{johnson_mimic-cxr_2019,
	title = {{MIMIC}-{CXR}, a de-identified publicly available database of chest radiographs with free-text reports},
	volume = {6},
	issn = {2052-4463},
	url = {https://www.nature.com/articles/s41597-019-0322-0},
	doi = {10.1038/s41597-019-0322-0},
	language = {en},
	number = {1},
	urldate = {2026-04-15},
	journal = {Scientific Data},
	author = {Johnson, Alistair E. W. and Pollard, Tom J. and Berkowitz, Seth J. and Greenbaum, Nathaniel R. and Lungren, Matthew P. and Deng, Chih-ying and Mark, Roger G. and Horng, Steven},
	month = dec,
	year = {2019},
	pages = {317},
}

@article{litjens_survey_2017,
	title = {A survey on deep learning in medical image analysis},
	volume = {42},
	issn = {13618415},
	url = {https://linkinghub.elsevier.com/retrieve/pii/S1361841517301135},
	doi = {10.1016/j.media.2017.07.005},
	language = {en},
	urldate = {2026-04-15},
	journal = {Medical Image Analysis},
	author = {Litjens, Geert and Kooi, Thijs and Bejnordi, Babak Ehteshami and Setio, Arnaud Arindra Adiyoso and Ciompi, Francesco and Ghafoorian, Mohsen and Van Der Laak, Jeroen A.W.M. and Van Ginneken, Bram and S{\'a}nchez, Clara I.},
	month = dec,
	year = {2017},
	pages = {60--88},
}

@article{chen_msa-net_2026,
	title = {{MSA}-{Net}: multi-scale attention-based {DenseNet} for multi-label chest {X}-ray image classification},
	volume = {113},
	issn = {17468094},
	shorttitle = {{MSA}-{Net}},
	url = {https://linkinghub.elsevier.com/retrieve/pii/S1746809425015800},
	doi = {10.1016/j.bspc.2025.109069},
	language = {en},
	urldate = {2026-04-15},
	journal = {Biomedical Signal Processing and Control},
	author = {Chen, Chao and Mat Isa, Nor Ashidi and Liu, Xin and Ding, Jurong and Lu, Ling},
	month = mar,
	year = {2026},
	pages = {109069},
}

@article{park_style-kd_2024,
	title = {Style-{KD}: {Class}-imbalanced medical image classification via style knowledge distillation},
	volume = {91},
	issn = {17468094},
	shorttitle = {Style-{KD}},
	url = {https://linkinghub.elsevier.com/retrieve/pii/S1746809423013617},
	doi = {10.1016/j.bspc.2023.105928},
	language = {en},
	urldate = {2026-04-15},
	journal = {Biomedical Signal Processing and Control},
	author = {Park, Inhyuk and Kim, Won Hwa and Ryu, Jongbin},
	month = may,
	year = {2024},
	pages = {105928},
}

@article{guan_multi-label_2020,
	title = {Multi-label chest {X}-ray image classification via category-wise residual attention learning},
	volume = {130},
	issn = {01678655},
	url = {https://linkinghub.elsevier.com/retrieve/pii/S0167865518308559},
	doi = {10.1016/j.patrec.2018.10.027},
	language = {en},
	urldate = {2026-04-15},
	journal = {Pattern Recognition Letters},
	author = {Guan, Qingji and Huang, Yaping},
	month = feb,
	year = {2020},
	pages = {259--266},
}

@article{ozturk_hydravit_2025,
	title = {{HydraViT}: {Adaptive} multi-branch transformer for multi-label disease classification from {Chest} {X}-ray images},
	volume = {100},
	issn = {17468094},
	shorttitle = {{HydraViT}},
	url = {https://linkinghub.elsevier.com/retrieve/pii/S1746809424010176},
	doi = {10.1016/j.bspc.2024.106959},
	language = {en},
	urldate = {2026-04-15},
	journal = {Biomedical Signal Processing and Control},
	author = {{\"O}zt{\"u}rk, {\c S}aban and Tural{\i}, M. Yi{\u g}it and {\c C}ukur, Tolga},
	month = feb,
	year = {2025},
	pages = {106959},
}

@misc{wang_bb-gcn_2023,
	title = {{BB}-{GCN}: {A} {Bi}-modal {Bridged} {Graph} {Convolutional} {Network} for {Multi}-label {Chest} {X}-{Ray} {Recognition}},
	copyright = {arXiv.org perpetual, non-exclusive license},
	shorttitle = {{BB}-{GCN}},
	url = {https://arxiv.org/abs/2302.11082},
	doi = {10.48550/ARXIV.2302.11082},
	urldate = {2026-04-15},
	publisher = {arXiv},
	author = {Wang, Guoli and Wang, Pingping and Cong, Jinyu and Liu, Kunmeng and Wei, Benzheng},
	year = {2023},
}

@inproceedings{ding_distilling_2025,
	address = {Xi'an, China},
	title = {Distilling {Label} {Co}-{Occurrence} {For} {Chest} {X}-{Ray} {Image} {Classification}},
	copyright = {https://doi.org/10.15223/policy-029},
	isbn = {979-8-3315-3626-8},
	url = {https://ieeexplore.ieee.org/document/11087133/},
	doi = {10.1109/ICSP65755.2025.11087133},
	urldate = {2026-04-15},
	booktitle = {2025 10th {International} {Conference} on {Intelligent} {Computing} and {Signal} {Processing} ({ICSP})},
	publisher = {IEEE},
	author = {Ding, Yunya},
	month = may,
	year = {2025},
	pages = {320--327},
}

@article{ward_global_2024,
	title = {Global {Burden} of {Disease} {Study} 2021 estimates: implications for health policy and research},
	volume = {403},
	issn = {01406736},
	shorttitle = {Global {Burden} of {Disease} {Study} 2021 estimates},
	url = {https://linkinghub.elsevier.com/retrieve/pii/S0140673624008122},
	doi = {10.1016/S0140-6736(24)00812-2},
	language = {en},
	number = {10440},
	urldate = {2026-04-15},
	journal = {The Lancet},
	author = {Ward, Zachary J and Goldie, Sue J},
	month = may,
	year = {2024},
	pages = {1958--1959},
}

@article{stacke_measuring_2021,
	title = {Measuring {Domain} {Shift} for {Deep} {Learning} in {Histopathology}},
	volume = {25},
	copyright = {https://ieeexplore.ieee.org/Xplorehelp/downloads/license-information/IEEE.html},
	issn = {2168-2194, 2168-2208},
	url = {https://ieeexplore.ieee.org/document/9234592/},
	doi = {10.1109/JBHI.2020.3032060},
	number = {2},
	urldate = {2026-04-15},
	journal = {IEEE Journal of Biomedical and Health Informatics},
	author = {Stacke, Karin and Eilertsen, Gabriel and Unger, Jonas and Lundstrom, Claes},
	month = feb,
	year = {2021},
	pages = {325--336},
}

@misc{chambon_roentgen_2022,
	title = {{RoentGen}: {Vision}-{Language} {Foundation} {Model} for {Chest} {X}-ray {Generation}},
	copyright = {arXiv.org perpetual, non-exclusive license},
	shorttitle = {{RoentGen}},
	url = {https://arxiv.org/abs/2211.12737},
	doi = {10.48550/ARXIV.2211.12737},
	urldate = {2026-04-15},
	publisher = {arXiv},
	author = {Chambon, Pierre and Bluethgen, Christian and Delbrouck, Jean-Benoit and Van der Sluijs, Rogier and Po{\l}acin, Ma{\l}gorzata and Chaves, Juan Manuel Zambrano and Abraham, Tanishq Mathew and Purohit, Shivanshu and Langlotz, Curtis P. and Chaudhari, Akshay},
	year = {2022},
}

@misc{han_deep_2015,
	title = {Deep {Compression}: {Compressing} {Deep} {Neural} {Networks} with {Pruning}, {Trained} {Quantization} and {Huffman} {Coding}},
	copyright = {arXiv.org perpetual, non-exclusive license},
	shorttitle = {Deep {Compression}},
	url = {https://arxiv.org/abs/1510.00149},
	doi = {10.48550/ARXIV.1510.00149},
	urldate = {2026-04-15},
	publisher = {arXiv},
	author = {Han, Song and Mao, Huizi and Dally, William J.},
	year = {2015},
}

@article{selvaraju_grad-cam_2016,
	title = {Grad-{CAM}: {Visual} {Explanations} from {Deep} {Networks} via {Gradient}-based {Localization}},
    journal = {arXiv preprint arXiv:1610.02391},
	copyright = {arXiv.org perpetual, non-exclusive license},
	shorttitle = {Grad-{CAM}},
	url = {https://arxiv.org/abs/1610.02391},
	doi = {10.48550/ARXIV.1610.02391},
	urldate = {2026-04-15},
	publisher = {arXiv},
	author = {Selvaraju, Ramprasaath R. and Cogswell, Michael and Das, Abhishek and Vedantam, Ramakrishna and Parikh, Devi and Batra, Dhruv},
	year = {2016},
}

@article{bustos_padchest_2020,
	title = {{PadChest}: {A} large chest x-ray image dataset with multi-label annotated reports},
	volume = {66},
	issn = {13618415},
	shorttitle = {{PadChest}},
	url = {https://linkinghub.elsevier.com/retrieve/pii/S1361841520301614},
	doi = {10.1016/j.media.2020.101797},
	language = {en},
	urldate = {2026-04-15},
	journal = {Medical Image Analysis},
	author = {Bustos, Aurelia and Pertusa, Antonio and Salinas, Jose-Maria and De La Iglesia-Vay{\'a}, Maria},
	month = dec,
	year = {2020},
	pages = {101797},
}

@misc{azizi_big_2021,
	title = {Big {Self}-{Supervised} {Models} {Advance} {Medical} {Image} {Classification}},
	copyright = {arXiv.org perpetual, non-exclusive license},
	url = {https://arxiv.org/abs/2101.05224},
	doi = {10.48550/ARXIV.2101.05224},
	urldate = {2026-04-15},
	publisher = {arXiv},
	author = {Azizi, Shekoofeh and Mustafa, Basil and Ryan, Fiona and Beaver, Zachary and Freyberg, Jan and Deaton, Jonathan and Loh, Aaron and Karthikesalingam, Alan and Kornblith, Simon and Chen, Ting and Natarajan, Vivek and Norouzi, Mohammad},
	year = {2021},
}

@article{seyyed-kalantari_underdiagnosis_2021,
	title = {Underdiagnosis bias of artificial intelligence algorithms applied to chest radiographs in under-served patient populations},
	volume = {27},
	issn = {1078-8956, 1546-170X},
	url = {https://www.nature.com/articles/s41591-021-01595-0},
	doi = {10.1038/s41591-021-01595-0},
	language = {en},
	number = {12},
	urldate = {2026-04-15},
	journal = {Nature Medicine},
	author = {Seyyed-Kalantari, Laleh and Zhang, Haoran and McDermott, Matthew B. A. and Chen, Irene Y. and Ghassemi, Marzyeh},
	month = dec,
	year = {2021},
	pages = {2176--2182},
}

@misc{angelopoulos_conformal_2022,
	title = {Conformal {Risk} {Control}},
	copyright = {arXiv.org perpetual, non-exclusive license},
	url = {https://arxiv.org/abs/2208.02814},
	doi = {10.48550/ARXIV.2208.02814},
	urldate = {2026-04-15},
	publisher = {arXiv},
	author = {Angelopoulos, Anastasios N. and Bates, Stephen and Fisch, Adam and Lei, Lihua and Schuster, Tal},
	year = {2022},
}

@inproceedings{chattopadhay_grad-cam_2018,
	address = {Lake Tahoe, NV},
	title = {Grad-{CAM}++: {Generalized} {Gradient}-{Based} {Visual} {Explanations} for {Deep} {Convolutional} {Networks}},
	isbn = {978-1-5386-4886-5},
	shorttitle = {Grad-{CAM}++},
	url = {https://ieeexplore.ieee.org/document/8354201/},
	doi = {10.1109/WACV.2018.00097},
	urldate = {2026-04-15},
	booktitle = {2018 {IEEE} {Winter} {Conference} on {Applications} of {Computer} {Vision} ({WACV})},
	publisher = {IEEE},
	author = {Chattopadhay, Aditya and Sarkar, Anirban and Howlader, Prantik and Balasubramanian, Vineeth N},
	month = mar,
	year = {2018},
	pages = {839--847},
}

@misc{springenberg_striving_2015,
	title = {Striving for {Simplicity}: {The} {All} {Convolutional} {Net}},
	shorttitle = {Striving for {Simplicity}},
	url = {http://arxiv.org/abs/1412.6806},
	doi = {10.48550/arXiv.1412.6806},
	urldate = {2026-04-15},
	publisher = {arXiv},
	author = {Springenberg, Jost Tobias and Dosovitskiy, Alexey and Brox, Thomas and Riedmiller, Martin},
	month = apr,
	year = {2015},
	note = {arXiv:1412.6806 [cs]},
}

@misc{sundararajan_axiomatic_2017,
	title = {Axiomatic {Attribution} for {Deep} {Networks}},
	copyright = {arXiv.org perpetual, non-exclusive license},
	url = {https://arxiv.org/abs/1703.01365},
	doi = {10.48550/ARXIV.1703.01365},
	urldate = {2026-04-15},
	publisher = {arXiv},
	author = {Sundararajan, Mukund and Taly, Ankur and Yan, Qiqi},
	year = {2017},
}

@misc{ribeiro_why_2016,
	title = {"{Why} {Should} {I} {Trust} {You}?": {Explaining} the {Predictions} of {Any} {Classifier}},
	copyright = {arXiv.org perpetual, non-exclusive license},
	shorttitle = {"{Why} {Should} {I} {Trust} {You}?},
	url = {https://arxiv.org/abs/1602.04938},
	doi = {10.48550/ARXIV.1602.04938},
	urldate = {2026-04-15},
	publisher = {arXiv},
	author = {Ribeiro, Marco Tulio and Singh, Sameer and Guestrin, Carlos},
	year = {2016},
}

@misc{zeiler_visualizing_2013,
	title = {Visualizing and {Understanding} {Convolutional} {Networks}},
	copyright = {arXiv.org perpetual, non-exclusive license},
	url = {https://arxiv.org/abs/1311.2901},
	doi = {10.48550/ARXIV.1311.2901},
	urldate = {2026-04-15},
	publisher = {arXiv},
	author = {Zeiler, Matthew D and Fergus, Rob},
	year = {2013},
}

@misc{lundberg_unified_2017,
	title = {A {Unified} {Approach} to {Interpreting} {Model} {Predictions}},
	copyright = {arXiv.org perpetual, non-exclusive license},
	url = {https://arxiv.org/abs/1705.07874},
	doi = {10.48550/ARXIV.1705.07874},
	urldate = {2026-04-15},
	publisher = {arXiv},
	author = {Lundberg, Scott and Lee, Su-In},
	year = {2017},
}

@misc{yun_cutmix_2019,
	title = {{CutMix}: {Regularization} {Strategy} to {Train} {Strong} {Classifiers} with {Localizable} {Features}},
	copyright = {arXiv.org perpetual, non-exclusive license},
	shorttitle = {{CutMix}},
	url = {https://arxiv.org/abs/1905.04899},
	doi = {10.48550/ARXIV.1905.04899},
	urldate = {2026-04-16},
	publisher = {arXiv},
	author = {Yun, Sangdoo and Han, Dongyoon and Oh, Seong Joon and Chun, Sanghyuk and Choe, Junsuk and Yoo, Youngjoon},
	year = {2019},
}

@misc{zhang_mixup_2018,
	title = {mixup: {Beyond} {Empirical} {Risk} {Minimization}},
	shorttitle = {mixup},
	url = {http://arxiv.org/abs/1710.09412},
	doi = {10.48550/arXiv.1710.09412},
	urldate = {2026-04-16},
	publisher = {arXiv},
	author = {Zhang, Hongyi and Cisse, Moustapha and Dauphin, Yann N. and Lopez-Paz, David},
	month = apr,
	year = {2018},
	note = {arXiv:1710.09412 [cs]},
}

@misc{grand-xray-slam-division-b,
    author = {Dhanjal, Guntas and Sammari, Salah and Ben Amor, Fathi and Aissa, Mariem},
    title = {Grand X-Ray Slam: Division B},
    year = {2025},
    howpublished = {\url{https://kaggle.com/competitions/grand-xray-slam-division-b}},
    note = {Kaggle}
}

\end{document}